\RequirePackage{fix-cm}
\documentclass[twocolumn]{svjour3}          
\usepackage{times}
\usepackage{epsfig}
\usepackage{graphicx}
\usepackage{amsmath}
\usepackage{amssymb,enumitem}

\usepackage{multirow}
\usepackage{tabls}
\usepackage{wrapfig}
\usepackage{hyperref}
\usepackage{amsmath,amssymb} 
\usepackage{amsmath,amssymb}
 \usepackage[ruled,vlined,linesnumbered,resetcount]{algorithm2e}
  \usepackage{booktabs}
 \usepackage[linesnumbered]{algorithm2e}
\usepackage{algorithmic}
\usepackage{caption}
\usepackage{subfigure}
\renewcommand{\algorithmicrequire}{ \textbf{Input:}} 
\renewcommand{\algorithmicensure}{ \textbf{Output:}} 
\usepackage[font=small,labelfont=bf]{caption}
   \usepackage{graphicx}
 \usepackage{booktabs}

\usepackage{epstopdf}
\usepackage{graphicx}
\usepackage{subfigure}
\usepackage{enumitem}
\usepackage{multirow}
\usepackage{tabls}
\usepackage{wrapfig}
\usepackage{hyperref}
\usepackage{amsmath,amssymb} 
\usepackage{amsmath,amssymb}
 \usepackage[ruled,vlined,linesnumbered,resetcount]{algorithm2e}
  \usepackage{booktabs}
 \usepackage[linesnumbered]{algorithm2e}
\usepackage{algorithmic}
\usepackage{caption}
\usepackage{subfigure}
\renewcommand{\algorithmicrequire}{ \textbf{Input:}} 
\renewcommand{\algorithmicensure}{ \textbf{Output:}} 
\usepackage[font=small,labelfont=bf]{caption}
   \usepackage{graphicx}
 \usepackage{booktabs}
\usepackage{epstopdf}
\usepackage{graphicx}
\usepackage{subfigure}
\usepackage{enumitem}
\smartqed  
\usepackage{graphicx}
\usepackage[T1]{fontenc}
\usepackage[utf8]{inputenc}

\usepackage{combelow}
\usepackage{newunicodechar}
\newunicodechar{ģ}{\cb{g}}
\newunicodechar{Ģ}{\cb{G}}
\journalname{International Journal of Computer Vision  }

\usepackage{gensymb} 

\begin{document}

\title{Synthesis of High-Quality Visible Faces from Polarimetric Thermal Faces using Generative Adversarial Networks\thanks{This work was supported by an ARO grant W911NF-16-1-0126.}
}


\author{He Zhang          \and Benjamin S. Riggan \and Shuowen  Hu \and Nathaniel J. Short \and
        Vishal M. Patel
}


\institute{He Zhang \at
              Department of ECE \\
              Rutgers, The State University of New Jersey\\
              94 Brett Road, Piscataway, NJ 08854\\
              \email{he.zhang92@rutgers.edu}           
                         \and
                         Benjamin S. Riggan
                         \at U.S. Army Research Laboratory\\
              \email{benjamin.s.riggan.civ@mail.mil}           
                         \and
                         Shuowen Hu
                         \at U.S. Army Research Laboratory\\
              \email{shuowen.hu.civ@mail.mil}           
                         \and
						Nathaniel J. Short \at
						Booz Allen Hamilton\\
              \email{short\_nathaniel@bah.com}           
                         \and
           Vishal M. Patel \at
              Department of ECE, Johns Hopkins University\\
               \email{vpatel36@jhu.edu}  
}

\date{Received: date / Accepted: date}

\maketitle

\begin{abstract}
The large domain discrepancy between faces captured
in polarimetric (or conventional) thermal and visible domain makes cross-domain face verification a highly challenging problem for human examiners as well as  computer vision algorithms. 
Previous approaches utilize either a two-step procedure (visible feature estimation and visible image reconstruction) or an input-level fusion technique, where different Stokes images are concatenated and used as a multi-channel input to synthesize the visible image given the corresponding polarimetric signatures.  Although these methods have yielded improvements, we argue that  input-level fusion alone may not be sufficient to realize the full potential of the available Stokes images.  We propose a Generative Adversarial Networks (GAN) based multi-stream feature-level fusion technique to synthesize high-quality visible images from prolarimetric thermal images.  The proposed network consists of a generator sub-network, constructed using an encoder-decoder network based on dense residual blocks, and a multi-scale discriminator sub-network.  The generator network is trained by optimizing an adversarial loss in addition to a  perceptual loss and an identity preserving loss to enable photo realistic generation of visible images while preserving discriminative characteristics. An {extended} dataset consisting of polarimetric thermal facial signatures of 111 subjects is also introduced.  Multiple experiments evaluated on different experimental protocols demonstrate that the proposed method achieves state-of-the-art performance. Code will be made available at  
\\https://github.com/hezhangsprinter.

\keywords{Face synthesis \and heterogeneous face recognition \and polarimetric data \and thermal face recognition \and deep learning \and generative adversarial networks.}

\end{abstract}

\section{Introduction}
\label{sec:intro}
Face is one of the most widely used biometrics for person recognition.  Various face recognition systems have been developed over the last two decades.  Recent advances in machine learning and computer vision methods have provided robust frameworks that achieve significant gains in performance of face recognition systems \cite{face_recognition1}, \cite{face_recognition2}, \cite{face_recognition3}, \cite{bodla2017deep}, \cite{xu2016learning}.  Deep learning methods, enabled by the vast improvements in processing hardware coupled with the ubiquity of face data, have led to significant improvements in face recognition accuracy, particularly in unconstrained face imagery \cite{DeepFace_SPM2018}, \cite{Chen2017_IJCV_face}, \cite{face_recognition4}.

Even though these methods are able to address many challenges and have even achieved human-expert level performance on challenging databases such as the low-resolution, pose variation and illumination variation to some extent ~\cite{DRGAN}, \cite{face_disentan_iccv2017}, \cite{face_recognition3}, \cite{face_recog_exp_fg2017}, \cite{DeepFace_SPM2018}, they are specifically designed for recognizing face images that are collected in the visible spectrum. Hence, they often do not perform well on the face images captured from other domains such as thermal \cite{face_btas2016}, \cite{face_ijcb2017}, \cite{face_thermal_hu}, \cite{facedata_cvprw17}, infrared \cite{Klare_NIR}, \cite{SWIR_TIFS}, \cite{wu2018disentangled} or millimeter wave \cite{mmW_TIFS}, \cite{mmW_ISBA2017} due to significant phenomenological differences as well as a lack of sufficient training data.

\begin{figure}[htp!]
	\centering
	\includegraphics[width=80mm]{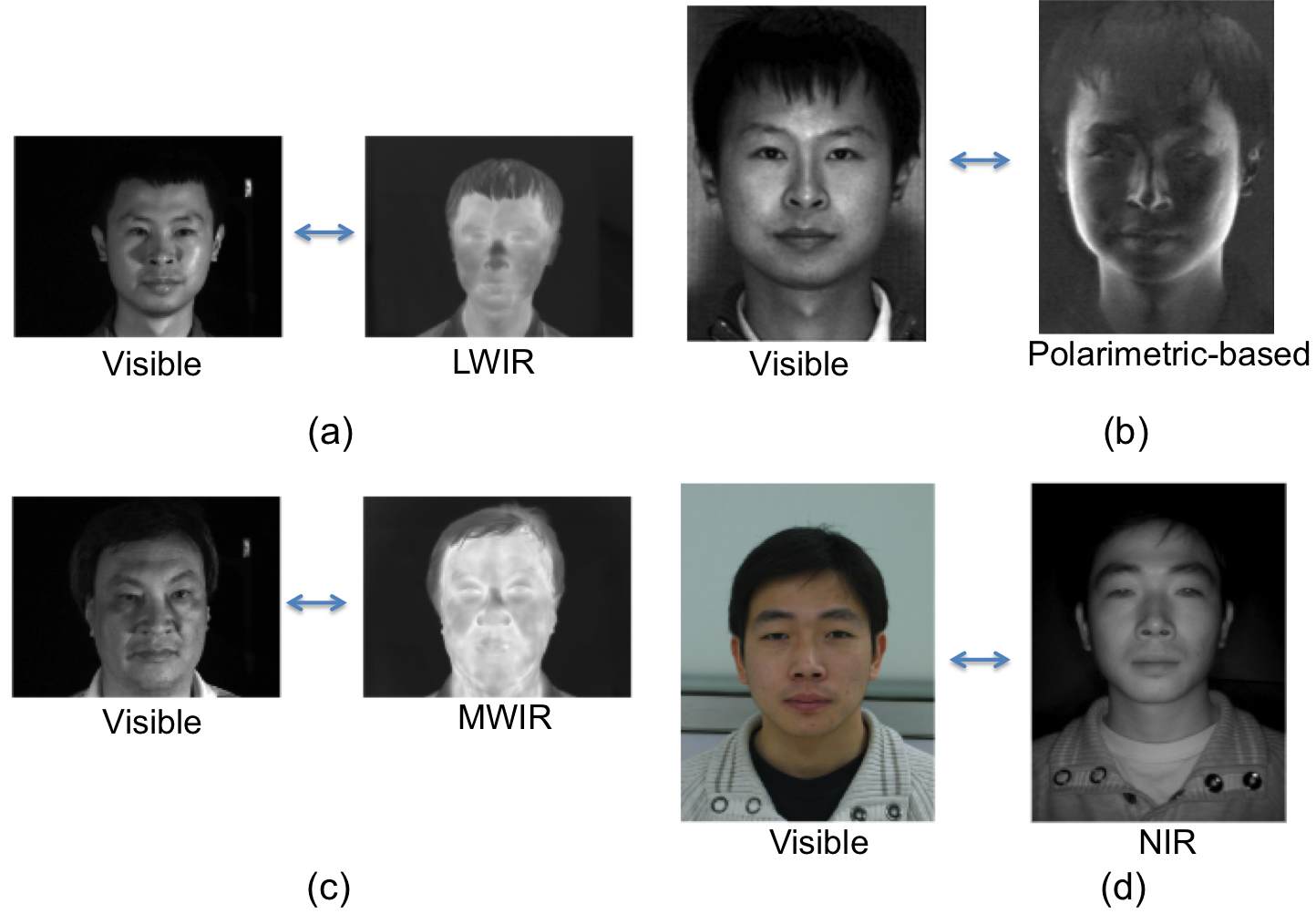}
	\caption{Examples of (a) visible-LWIR pair \cite{Chris_IEEEAccess}, (b) visible-polarimetric pair \cite{short2015improving}, (c) visible-MWIR pair \cite{Chris_IEEEAccess}, and (d) visible-NIR pair \cite{Chris_IEEEAccess}.}
	\label{fig:modalities}
\end{figure}

Thermal imaging has been proposed for night-time and low-light face recognition when external illumination is not practical due to various collection considerations. The infrared spectrum can be divided into a reflection dominated region consisting of the near infrared (NIR) and shortwave infrared (SWIR) bands, and an emission dominated thermal region consisting of the midwave infrared (MWIR) and longwave infrared (LWIR) bands \cite{face_thermal_wacv2016}.  In particular, recent works have been proposed to use the polarization-state information of thermal emissions to enhance the performance of thermal face recognition \cite{facedata_cvprw17}, \cite{face_btas2016}, \cite{short2015improving}, \cite{face_ijcb2017}.  It has been shown that polarimetric-thermal images capture geometric and textural details of faces that are not present in the conventional thermal facial imagery \cite{short2015improving}.    As a result, the use of polarization-state information can improves cross-spectrum recognition performance over using intensity-only information from conventional thermal imagers.

Thermal face imagery, which can be acquired passively at night, but are currently  not  maintained in biometric-enabled watch lists, must be compared with visible-light face images for interoperability with existing biometric face databases.   Distributional/domain differences between thermal and visible images makes thermal-to-visible face recognition very challenging (see Figure~\ref{fig:modalities}).  Various methods have been developed in the literature to bridge this gap for cross-domain (i.e., heterogeneous) face recognition \cite{face_thermal_bmvc2015,face_thermal_hu,face_thermal_ijcv2017,face_thermal_wacv2016,face_thermal_icb2018, bodla2017deep}.  In particular, methods that synthesize visible faces from thermal facial signatures have gained  traction in recent years \cite{face_btas2016}, \cite{face_ijcb2017}.  One of the advantages of face synthesis is that once the face images are synthesized in the visible domain, any off-the-shelf face matching algorithm can be used to match the synthesized image to the gallery of visible images.

A polarimetric signature/image is defined here as consisting of three Stokes images as its three channels, analogous to the RGB channels in visible color imagery. Previous approaches utilize either a two-step procedure (visible feature estimation and visible image reconstruction) \cite{face_btas2016} or a fusion technique where different Stokes images are concatenated and used as a multi-channel input \cite{face_ijcb2017} to synthesize the visible image.  Though these methods are able to effectively synthesize photo-realistic visible face images, the results are still far from optimal. One possible reason lies in that these methods concatenate the Stokes images into a single input sample  without any additional attempts to capture multi-channel information inherently present in the different Stokes (modalities) images from the thermal infrared band. Hence, in order to efficiently leverage the multi-modal information provided by the polarimetric thermal images, we propose a novel multi-stream feature-level fusion method for synthesizing visible images from thermal domain using recently proposed Generative Adversarial Networks \cite{GAN}.

\begin{figure*}[htp!]
	\centering
	\includegraphics[width=1\textwidth]{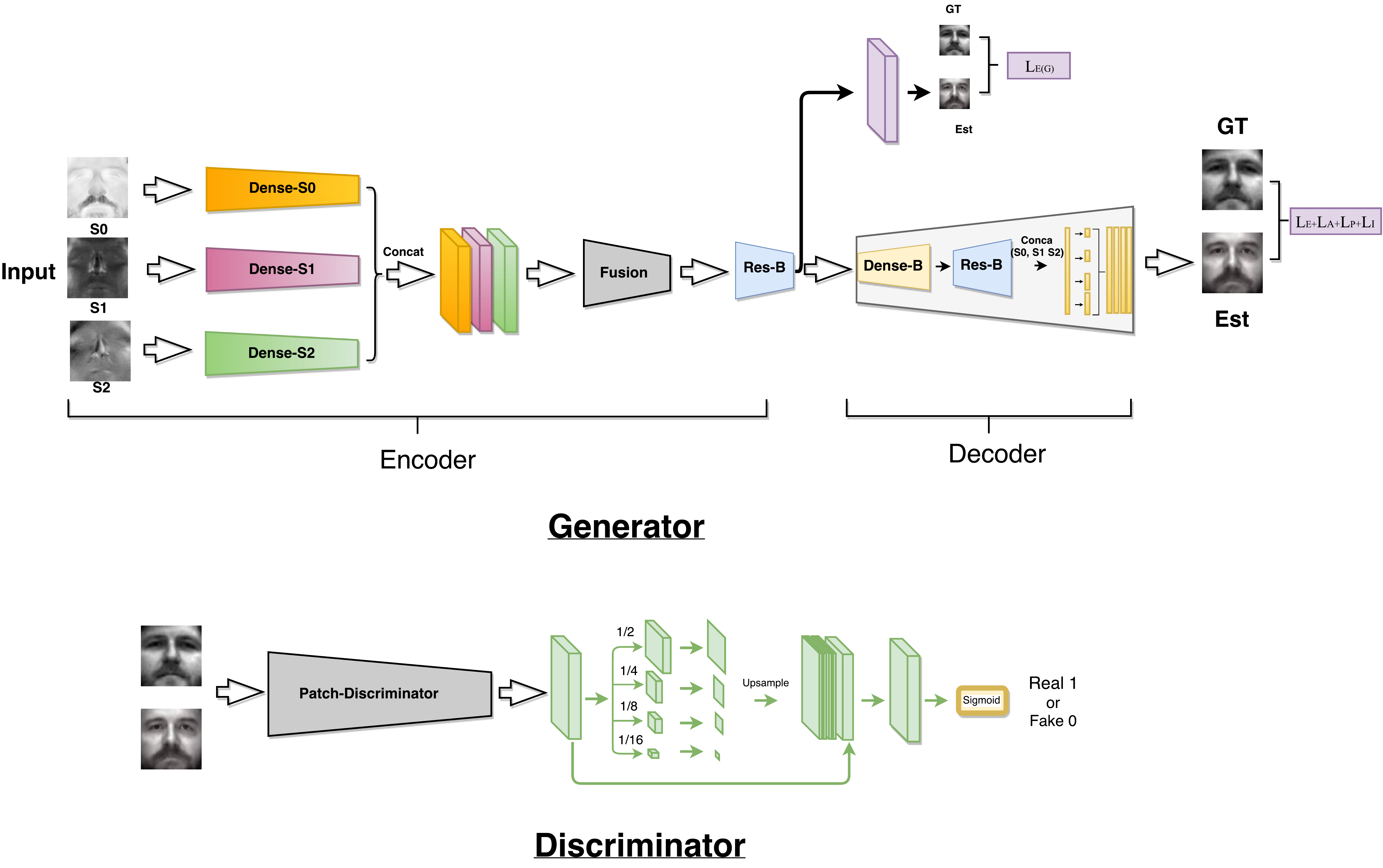}
	\vskip 0pt  \caption{An overview of the proposed GAN-based multi-stream encoder-decoder network. The generator contains a multi-stream feature-level fusion encoder-decoder network. In addition, a deep-guided subnet is stacked at the end of the encoding part.  The discriminator is composed of a multi-scale patch-discriminator  structure.}
	\label{fig:overview}
\end{figure*}

\begin{figure*}[t]
	\centering
	\begin{minipage}{.15\textwidth}
		\centering
		\includegraphics[width=1\textwidth]{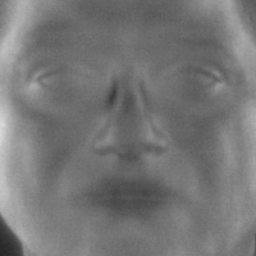}
		\captionsetup{labelformat=empty}
		\captionsetup{justification=centering}
	\end{minipage}
	\begin{minipage}{.15\textwidth}
		\centering
		\includegraphics[width=1\textwidth]{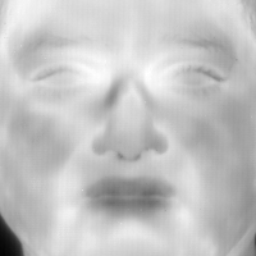}
		\captionsetup{labelformat=empty}
		\captionsetup{justification=centering}
	\end{minipage}
	\begin{minipage}{.15\textwidth}
		\centering
		\includegraphics[width=1\textwidth]{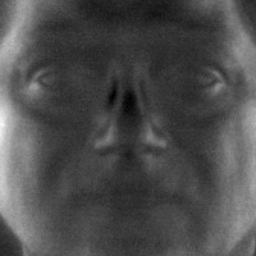}
		\captionsetup{labelformat=empty}
		\captionsetup{justification=centering}
	\end{minipage}
	\begin{minipage}{.15\textwidth}
		\centering
		\includegraphics[width=1\textwidth]{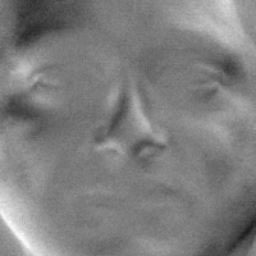}
		\captionsetup{labelformat=empty}
		\captionsetup{justification=centering}
	\end{minipage}
	\begin{minipage}{.15\textwidth}
		\centering
		\includegraphics[width=1\textwidth]{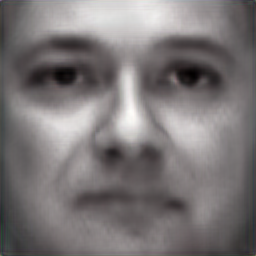}
		\captionsetup{labelformat=empty}
		\captionsetup{justification=centering}
	\end{minipage}
	\begin{minipage}{.15\textwidth}
		\centering
		\includegraphics[width=1\textwidth]{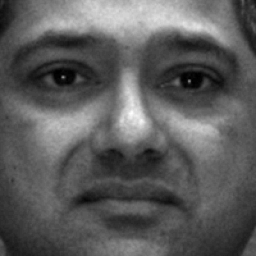}
		\captionsetup{labelformat=empty}
		\captionsetup{justification=centering}
	\end{minipage}\\
	\begin{minipage}{.15\textwidth}
		\centering
		\includegraphics[width=1\textwidth]{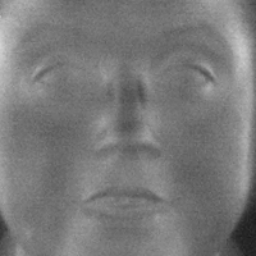}
		\captionsetup{labelformat=empty}
		\captionsetup{justification=centering}
	\end{minipage}
	\begin{minipage}{.15\textwidth}
		\centering
		\includegraphics[width=1\textwidth]{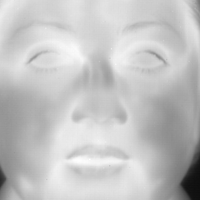}
		\captionsetup{labelformat=empty}
		\captionsetup{justification=centering}
	\end{minipage}
	\begin{minipage}{.15\textwidth}
		\centering
		\includegraphics[width=1\textwidth]{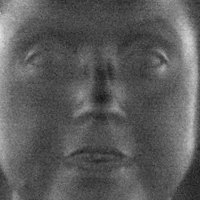}
		\captionsetup{labelformat=empty}
		\captionsetup{justification=centering}
	\end{minipage}
	\begin{minipage}{.15\textwidth}
		\centering
		\includegraphics[width=1\textwidth]{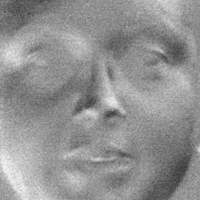}
		\captionsetup{labelformat=empty}
		\captionsetup{justification=centering}
	\end{minipage}
	\begin{minipage}{.15\textwidth}
		\centering
		\includegraphics[width=1\textwidth]{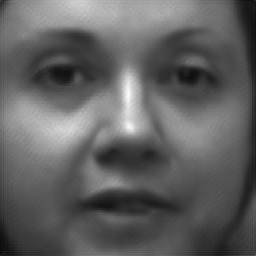}
		\captionsetup{labelformat=empty}
		\captionsetup{justification=centering}
	\end{minipage}
	\begin{minipage}{.15\textwidth}
		\centering
		\includegraphics[width=1\textwidth]{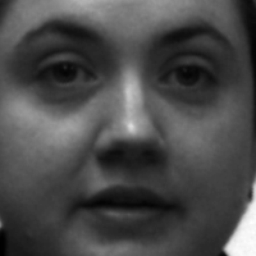}
		\captionsetup{labelformat=empty}
		\captionsetup{justification=centering}
	\end{minipage}\\
	\centering
	\begin{minipage}{.15\textwidth}
		\centering
		\includegraphics[width=1\textwidth]{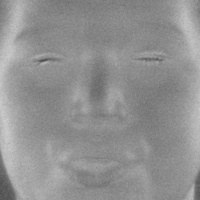}
		\captionsetup{labelformat=empty}
		\captionsetup{justification=centering}
		\caption*{Polar}
	\end{minipage}
	\begin{minipage}{.15\textwidth}
		\centering
		\includegraphics[width=1\textwidth]{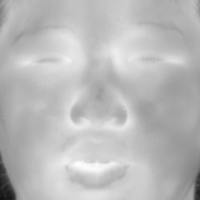}
		\captionsetup{labelformat=empty}
		\captionsetup{justification=centering}
		\caption*{$S_0$}
	\end{minipage}
	\begin{minipage}{.15\textwidth}
		\centering
		\includegraphics[width=1\textwidth]{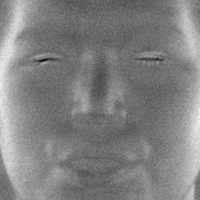}
		\captionsetup{labelformat=empty}
		\captionsetup{justification=centering}
		\caption*{$S_1$}
	\end{minipage}
	\begin{minipage}{.15\textwidth}
		\centering
		\includegraphics[width=1\textwidth]{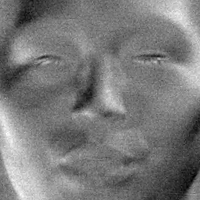}
		\captionsetup{labelformat=empty}
		\captionsetup{justification=centering}
		\caption*{$S_2$}
	\end{minipage}
	\begin{minipage}{.15\textwidth}
		\centering
		\includegraphics[width=1\textwidth]{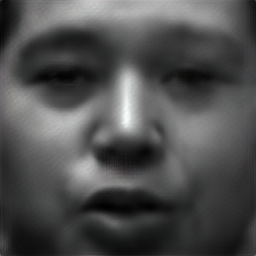}
		\captionsetup{labelformat=empty}
		\captionsetup{justification=centering}
		\caption*{Proposed}
	\end{minipage}
	\begin{minipage}{.15\textwidth}
		\centering
		\includegraphics[width=1\textwidth]{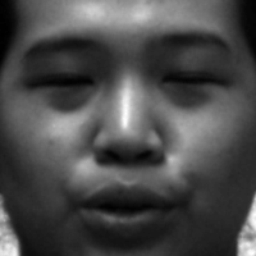}
		\captionsetup{labelformat=empty}
		\captionsetup{justification=centering}
		\caption*{Target}
	\end{minipage}\\	
	\vskip -8pt
	\caption{Sample results of the proposed method. (a) Input Polarimetric image.  (b) Input $S_0$ image.  (c) Input $S_1$ image.  (d) Input $S_2$ image (e) Results from the proposed method, and (f) Target image.}  \label{fig:over_image}
\end{figure*}

The proposed GAN-based network consists of a generator, a discriminator sub-network and a deep guided sub-network (see Figure~\ref{fig:overview}). The generator is composed of a multi-stream encoder-decoder network based on dense-residual blocks, the discriminator is designed to capture features at multiple-scales for discrimination and the deep guided sub-net aims to guarantee that  the encoded features contain geometric and texture information to recover the visible face.  To  further  enhance  the network's  performance,  it  is  guided  by  perceptual  loss  and an identity  preserving  loss  in  addition  to  adversarial  loss.  Once the face images are synthesized, any off-the-shelf face recognition and verification networks trained on the visible-only face data can be used for matching.  Figure \ref{fig:over_image} illustrates the differences between visible and polarimetric thermal images. In addition, this figure also presents the photo-realistic and identity-preserving results obtained from our proposed method.

In addition to developing a novel face synthesis network, we also collected an extended dataset containing of visible and polarimetric facial signatures from 111 subjects.  A subset of this dataset consisting data from 60 subjects was described in \cite{facedata_cvprw17}.  This extended  polarimetric thermal facial dataset is available to computer vision and  biometrics researchers to facilitate the development of cross-spectrum and multi-spectrum face recognition algorithms.

To summarize, this paper makes the following contributions. 
\begin{enumerate}
\item A novel face synthesis framework based on GAN is proposed which consists of a multi-stream generator and multi-scale discriminator.  

\item To embed the identity information into the objective function and make sure that the synthesized face images are photo-realistic, a refined loss function is proposed for training the network.

\item An extended dataset consisting of visible and polarimetric data from 111 subjects is collected.

\item  Detailed experiments are conducted to demonstrate improvements  in  the  synthesis  results.  Further,  three ablation
studies  are  conducted  to  verify  the  effectiveness  of
iterative synthesis and various loss function.
\end{enumerate}

Compared to the previous approaches for polarimetric thermal to visible face synthesis, this is a completely new work and one of the first approaches that proposes to use GANs for synthesizing high-quality visible faces from polarimetric thermal faces. Our previous work \cite{face_ijcb2017} also addresses the same problem but there are several notable differences: 1) The newly proposed method includes a novel multi-stream densely-connected network to transfer the polarimetric thermal facial signatures into the visible domain.   
2) Feature-leavel fusion of different Stokes images is explored in this work to demonstrate the  effectiveness of leveraging multiple polarimetric modalities for visible face synthesis.
3) We introduce an extended dataset that includes 111 subjects.  
4) A novel multi-scale discriminator is introduced that leverages the  information from different scales to decide whether the given image is real or fake. 

The paper is organized as follows. In Section~\ref{sec:background}, we review  related works and give a brief background on GANs and polarimetric thermal imaging.    Details of the proposed multi-stream feature-level fusion method are discussed in Section~\ref{sec:proposed_methods}.  In Section~\ref{sec:database}, we describe the extended polarimetric thermal face dataset.  Experiments and results are presented in Section~\ref{sec:experiment} and Section~\ref{sec:con} concludes the paper with a brief summary and discussion.

\section{Background and Related Work}
\label{sec:background}

In this section, we give a brief overview of polarimetric thermal imaging and GANs. Then, we  review some related works on heterogeneous face recognition algorithms.

\subsection{Polarimetric Thermal Imaging}
\label{sssec:polar}
Polarimetric thermal imaging uses advanced materials and sensor technology to measure the polarization state of light. While traditional imaging exploits the intensity of light, polarimetric imaging exploits the orientation of its oscillations. Natural visible light exhibits no preferred polarization state. If natural light is either transmitted across a boundary from one medium to another, or is reflected by the boundary (i.e., the material is opaque), a preferential polarization state (usually linear)  may occur.

This induced polarization change is a directional quantity and is a function of the angle between the surface normal and the transmitted/reflected ray. For example, unpolarized sunlight reflecting off an air-water interface 
results in an induced linear polarization state that is orthogonal to the 
plane of reflection, as defined by the surface normal and the reflected ray. A similar phenomena exists when considering light energy in the ``thermal"  infrared (IR) part of the spectrum, e.g., MidIR (3-5$\mu$m) and/or LWIR (8-12$\mu$m). For induced polarization in the thermal IR, the radiation is treated as either  emitted and/or reflected from a surface boundary. It is this interaction at  the boundary that results in an induced net linear polarization state, similar to situation seen for visible light.  By capturing this thermal radiance using  an IR polarimetric camera, one can exploit the additional polarization based  information and reconstruct a 3D surface from a 2D polarimetric image.

Polarimetric imaging sensors capture polarization-state information through optical filtering of light at different polarizations.  This is traditionally done using a rotating element~\cite{tyo2006review} (i.e., division of time), but other approaches exist, such as micro-grid polarizers~\cite{tyo2006review} (i.e., division of focal plane array). In essence, polarization-state information is captured at four orientations, $I_0$, $I_{90}$, $I_{45}$, and $I_{135}$.  The $I_0$ and $I_{90}$ measurements represent horizontal and vertical polarized light and $I_{45}$ and $I_{135}$ capture diagonally polarized light.  A stack of 2-D images captured using a polarimeter is represented by Stokes images, as defined in~\cite{gurton2014enhanced}, which contain geometric and textural features, such as edges of the nose and mouth as well as wrinkles.   These Stokes images are illustrated in Figure \ref{fig:over_image} for three subjects with corresponding visible-spectrum facial signatures.  The $S_0$ image is a total intensity polarimetric image and is representative of what a conventional thermal imager (i.e., without linear polarizer) would capture. $S_1$, and $S_2$  illustrate the additional details provided by polarimetric imaging.  In this paper, we refer to \emph{Polar} as the three channel polarimetric image with $S_0$, $S_1$ and $S_2$ as the three channels.

\subsection{Generative Adversarial Networks }
 \label{sec:gan}
Generative Adversarial Networks were first proposed by Goodfellow et al. in \cite{GAN} to synthesize realistic images by effectively learning the distribution of the training images. The authors adopted a game theoretic min-max optimization framework to simultaneously train two models: a generative model, $G$, and a discriminative model, $D$.
Initially, the success of GANs was limited as they were known to be unstable to train, often resulting in artifacts in the synthesized images. Radford \textit{et al.} in \cite{GAN_unsupervised} proposed Deep Convolutional GANs (DCGANs) to address the issue of instability by including a set of constraints on their network design. Another limiting issue in GANs is that, there is no control on the modes of data being synthesized by the generator in case of these unconditioned generative models. Mirza \textit{et al.} \cite{mirza2014conditional} incorporated additional conditional information in the model, which resulted in effective learning of the generator. The use of conditioning variables for augmenting side information not only increased the stability in learning but also improved the descriptive power of the generator $G$ \cite{karacan2016learning}. Recently, researchers have explored various aspects of GANs such as training improvements \cite{salimans2016improved} and use of task specific cost function \cite{creswell2016task}. Also, an alternative viewpoint for the discriminator function is explored by Zhao \textit{et al.} \cite{zhao2016energy} where they deviate from the traditional probabilistic interpretation of the discriminator model.  The objective function of a conditional  GAN is defined as follows
\begin{equation}\label{eq:conditional GAN loss}
\begin{split}
L_{cGAN}(G,D) = E_{x,y \sim Pdata (x,y)}[\log D(x,y)]+ \\
E_{x\sim Pdata(x),z\sim p_{z}(z)}[\log(1-D(x,G(x,z)))],
\end{split}
\end{equation}
where $y$, the output image, and $x$, the observed image, are sampled from distribution $Pdata (x,y)$ and they are distinguished by the discriminator, $D$. While for the generated fake $G(x,z)$ sampled from distributions $x\sim Pdata(x),z\sim p_{z}(z)$ would like to fool $D$. 
 
 The success of GANs in synthesizing realistic images has led  researchers to explore the GAN framework for numerous applications such as data augmentation \cite{peng2018jointly}, zero-shot learning \cite{zhu2017imagine}, image inpainting \cite{yu2018generative}, image dehazing \cite{xustrong,zhang2018densely,yang2018towards}, text-to-image translation \cite{xu2017attngan}, image-to-image translation \cite{pix2pix,zhang2018translating,perera2017in2i}, texture synthesis \cite{jetchev2016texture}, crowd-counting \cite{gan_crowd} and generating outdoor scenes from attributes \cite{karacan2016learning}. Isola \textit{et al.} \cite{pix2pix} proposed a general method for image-to-image translation using conditional adversar`ial networks. Apart from learning a mapping function, they argue that the network also learns a loss function, eliminating the need for specifying or designing a task specific loss function. Karacan \textit{et al.} in \cite{karacan2016learning} proposed a deep GAN conditioned on semantic layout and scene attributes to synthesize realistic outdoor scene images under different conditions. Recently, Jetchev \textit{et al.} \cite{jetchev2016texture} proposed spatial GANs for texture synthesis. Different from traditional GANs, their input noise distribution constitutes a whole spatial tensor instead of a vector, thus enabling them to create architectures more suitable for texture synthesis.
 
 \subsection{Heterogeneous Face Recognition}

Recently, there has been a growing number of approaches that bridge existing modality gaps in order to perform heterogeneous face recognition.  These approaches focused on various scenarios of heterogeneous face recognition such as  infrared-to-visible \cite{face_nir_cvpr17,face_nir_casia2017,song2017adversarial}, thermal-to-visible \cite{face_thermal_bmvc2015,face_thermal_wacv2016,face_ijcb2017,face_thermal_icb2018,Klare_NIR}, and sketch-to-visible \cite{face_sketch_arxiv_2018,face_sketch_TNN2016} \cite{Lidan_Sketch_FG18}.  Fundamentally, each approach seeks to either find a common latent subspace in which corresponding faces from each domain are ``close'' in terms of some distance and non-corresponding faces are ``far.'' or synthesize  photo-realistic  visible face given its corresponding input modality.
 
Klare and Jain~\cite{face_recogniton_heter1} proposed an approach using kernel prototype similarities, where after geometric normalization and image filtering (e.g., Difference of Gaussian, Center-Surround Divisive Normalization ~\cite{face_recogniton_prepro}, and Gaussian) and local features extraction (e.g., multi-scale local binary patterns, or MLBP, and scale invariant feature transform, or SIFT), the intra-domain kernel similarities are computed between source (or target) domain images and all training images from the source (or target) domain.  These intra-domain kernel similarity, which are computed using the cosine kernel, provide relational vectors for source and target domain imagery to be compared, where the main idea is that the kernel similarity between two source domain images should be similar to the kernel similarity between two corresponding target domain images.
 
Yi \emph{et al.}~\cite{face_recognition_hfr} leverage the use of multi-modal Restricted Boltzmann Machines (RBMs)~\cite{restricted_boltzmann_machines} to learn a shared representation, and for NIR-to-visible face recognition.  Here, they learn shared representation using the multi-modal RBMs locally for each patch.  However, since heterogeneity is only addressed locally, they further reduce the modality gap by performing Hetra-component analysis (HCA)~\cite{face_nirdatasets} for the holistic image representation.  Hetero-component analysis is based on the theory that most of the appearance differences between imaging modalities are captured in the top eigenvectors.  Therefore, a common representation is given by removing the effects from the top eigenvectors.  This was shown to achieve excellent performance for NIR-to-visible face recognition.  However, it is unclear how well this would work for an emissive infrared band, such LWIR, where facial signatures are very different than in visible or NIR bands due to phenomenology.
 
Riggan \emph{et al.}~\cite{Chris_IEEEAccess} proposed coupled auto-associative network for learning common representation between thermal and visible face images.  The authors optimize two sparse auto-encoders jointly, such that (1) information within each modality is preserved and (2) inter-domain representations are similar for corresponding images.  Although this approach demonstrated some success and robustness, the constraint to preserve information for the source domain is not a necessary condition as long as discriminability is maintained when learning common representation.
 
Hu \emph{et al.}~\cite{face_thermal_hu} applied a one-versus-all framework using partial least squares classifiers on Histogram of Oriented Gradients (HOG) features.  For each classifier, they introduce the concept of adding cross-domain negative samples (i.e., thermal samples from a different subject) for added robustness.  Later, Riggan \emph{et al.}~\cite{face_thermal_wacv2016} proposed the using of a coupled neural network and a discriminative classifier for enhance conventional thermal-to-visible face recognition and polarimetric thermal-to-visible framework.
 
While the described methods individually introduce custom approaches for reducing the modality gaps between two imaging domains, there are two fundamental concerns: (1) how to adjudicate matches when returned to an analyst, and (2) how to leverage recent advances in visible spectrum face recognition.  Therefore, Riggan \emph{et al.}~\cite{face_btas2016} proposed a way to synthesize a visible image from both conventional thermal and polarimetric thermal.  This approach used a convolution neural network to extract features from a conventional or polarimetric thermal image and then mapped those features to a corresponding visible representation using a deep perceptual mapping~\cite{ivertingcnn_vgg}, where this representation in inverted back to the imaging domain using the forward convolutional neural network model.  One potential concern is the piece-wise nature of this synthesis method.  Later, built on the success of generative adversarial networks \cite{GAN}, Zhang \emph{et al.}~\cite{face_ijcb2017} improved the synthesis results by proposing an end-to-end conditional generative adversarial network (CGAN) approach, which is optimized via a  newly introduced identity-preserving loss, to synthesis a corresponding visible image given a thermal image.  This approach demonstrated results that were photo-realistic and discriminative.

\section{Proposed Method}
\label{sec:proposed_methods}

As discussed earlier, a polarimetric sample consists of three different Stokes images ($S_0$, $S_1$ and $S_2$), where $S_0$ represents the conventional thermal image and $S_1$ and $S_2$ represent the horizontal/vertical and diagonal polarization-state information, respectively. Unlike traditional three-channel RGB images where each channel contains different spectral information, the $S_0$, $S_1$, $S_2$ images contain different geometric and texture information. For example, as shown in the first row of  Figure \ref{fig:over_image}, $S_0$ is able to capture the mustache information, which is not captured in $S_1$ and $S_2$.  On the other hand, $S_0$ does not capture some of the other texture and geometric details such as wrinkles and the shape of the mouth, which are well-preserved in $S_1$ and $S_2$.  In other words, the Stokes images individually capture different facial features and when combined together they provide complementary information. Hence, it is important to fully utilize the information from  all three Stokes images to  effectively synthesize a visible face image.

Previous methods have attempted to utilize this information by exploiting input level fusion, where three  Stokes images are concatenated together as a three-channel input \cite{face_btas2016,face_ijcb2017}.  Even though the three-channel concatenation in the input level is able to generate better visible face results by bringing in the geometric and texture differences preserved in these three modalities as compared with using just a single Stokes image as input (eg. $S_0$), the results are still far from optimal \cite{face_ijcb2017}. A potential reason is that input level fusion or mere concatenation of different Stokes images may not be sufficient enough to exploit the different geometric and texture information present in these modalities. \footnote{Input level fusion can be regarded as an extreme case for low-level feature fusion, where low-level  features (from shallow layers) often preserve edge information rather than semantic mid-level or high-level class-specific information \cite{cnn_visualizing}.}.  To efficiently address this problems and generate better photo-realistic visible face images, a multi-stream feature-level fusion structure is proposed in this paper.  Specifically,  different encoder structures are leveraged to encode each Stokes image separately and then the embedded features from  each encoder are fused together via a fusion block for further visible face reconstruction (i.e. decoding).

Synthesizing photo-realistic visible images from polarimetric images (or even any single Stokes image) is an extremely challenging problem due to information differences caused by phenomenology between polarimetric thermal images and visible images. As shown in Figure \ref{fig:over_image}, polarimetric thermal images fail to capture fine details such as edges and gradients as compared to visible images. Due to the absence of these sharp details in the polarimetric images, synthesizing visible images from them requires joint modeling of the images from these two domains. To efficiently leverage the training samples and  guarantee better convergence with less gradient vanishing for such joint modeling, a novel dense residual structure is proposed in this paper. Furthermore, a multi-scale patch-discriminator is utilized to classify between real and synthesized images at multiple scales. By performing the discrimination at  multiple scales, we are able to effectively leverage contextual information in the input image, resulting in better high-frequency details in the reconstructed image.

To summarize, we propose a multi-stream feature-level fusion GAN structure (see Figure~\ref{fig:overview}) which consists of the following components: 
\begin{enumerate}[nolistsep]
\item[(1)] Multi-stream densely-connected encoder. 
\item[(2)] Deep guidance sub-network.
\item[(3)] Single-stream dense residual decoder.
\item[(4)] Multi-scale discriminator.
\end{enumerate}
In what follows, we describe these components in detail.

\subsection{Multi-stream Feature-level Fusion Generator}
The proposed feature-level fusion method is inspired by the  face dis-entangled representation work proposed by Peng \emph{et al.} and Tran \emph{et al.} in \cite{face_disentan_iccv2017,DRGAN,face_dis_eccv16}, where the encoded feature representations are explicitly disentangled into separate parts  representing different facial priors such as identity, pose and gender. Rather than leveraging the supervised label information to enforce the disentangling factor in the embedded features, each encoder structure in the proposed method inherently learns to characterize different geometric and texture information that is captured in the Stokes images. This information is then combined with a residual block-based fusion network, followed by a decoder network, consisting of a dense network and a residual network, to reconstruct visible domain faces from the fused feature maps. Furthermore, a deep-guided sub-network is leveraged at the end of the encoding part to ensure that the encoded features preserve geometric and texture information.\\

\begin{figure}[htp!]
	\centering
	\includegraphics[width=0.5\textwidth]{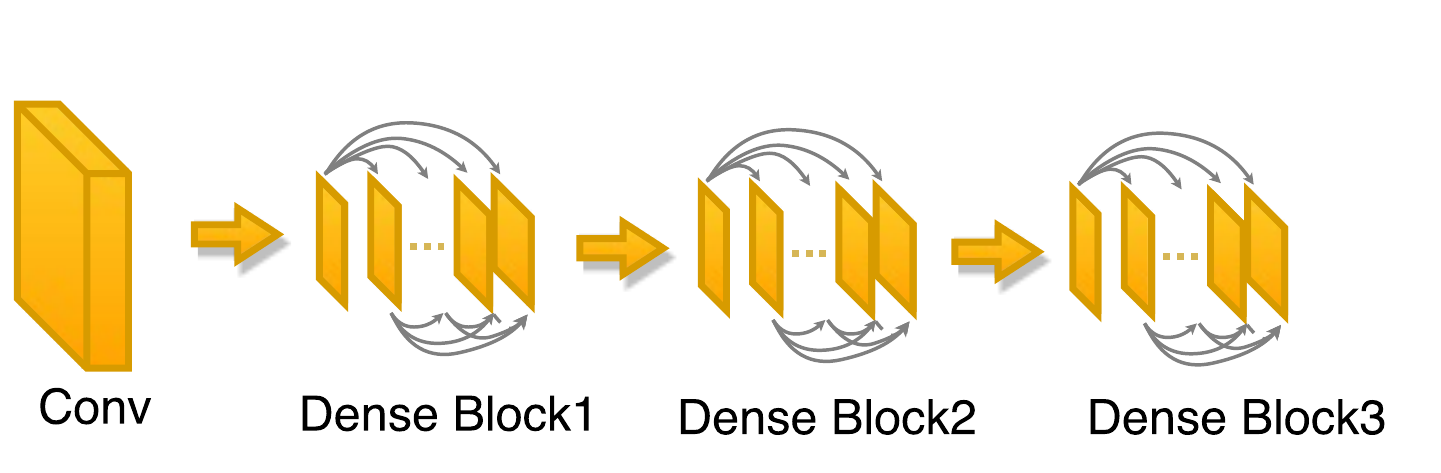}
	\vskip -5pt  \caption{Overview of a single stream in the multi-stream densely-connected encoding part.}
	\label{fig:overview_encoder}
\end{figure}

\noindent {\bf{Multi-stream Densely-connected Encoding.}}
The encoder consists of three streams of sub-networks, with each sub-network having the same structure. \footnote{Weights are not shared among each stream.}. Each stream processes a particular input Stokes image. Basically, each stream is composed of a convolutional layer with kernel size 4, stride 2  and zero-padding 1, rectified linear unit (ReLU) and a 2x2 max-pooling operator at the front followed by three level dense-blocks \cite{dense_net} \footnote{Feature map size (width and height) in each level is same.  }. Each layer $\mathbf{D}_j$ in a dense block can be represented as 
\begin{equation}\label{eq:single_stream}
\mathbf{D}_j=T({cat}[D_1, D_2,..., D_{j-1}]),
\end{equation}
where $T(\cdot)$ indicates the combination of Batch Normalization (BN) \cite{batch_norm}, rectified linear unit (ReLU) and  Convolution operator. Figure~\ref{fig:overview_encoder} gives an overview of a single stream in the multi-stream densely-connected encoding.

There are three levels of densely connected blocks where each separate level contains 12, 16, or 24 dense-blocks. In addition, there exist one transition down layer  which is composed of a 1$\times$1 convolution layer  followed by $2\times2$ average pooling  between two contiguous dense blocks.   All the parameters in each stream in the encoder can be initialized using the pre-trained Dense-net 121 structure \cite{dense_net}.  Each dense-block contains a $3\times3$ convolution layer with ReLU and Batch normalization. In addition, the output of each dense block is concatenated with the input of the corresponding dense block.  Once we calculated features  from all three streams, we concatenate together from all three branches along the depth (channel) dimension.   Feature maps from each of the three streams are of size $C \times H\times W$. These feature maps are concatenated and are forwarded to the residual-fusion block, which consists of a res-block with $1\times1$ convolution layer. Then, the output of the residual-fusion block is regarded as the input  for two different branches.  To guarantee that the learned features contain geometric and textural facial information, a deep guidance sub-network \cite{deep_supervision} is introduced at the end of the encoding part as one branch. The deep guided sub-network is part of the network that is branching out from the  end of the encoder. This sub-network is composed of a 1$\times$1 convolution layer followed by the non-linear function, Tanh.  Hence, the output of the guided sub-network  will be a three-channel RGB  image with size $16\times16$ if the input size is $256\times256$. In addition, the decoder is regarded as another branch discussed below. \\

\begin{figure}[htp!]
	\centering
	\includegraphics[width=0.46\textwidth]{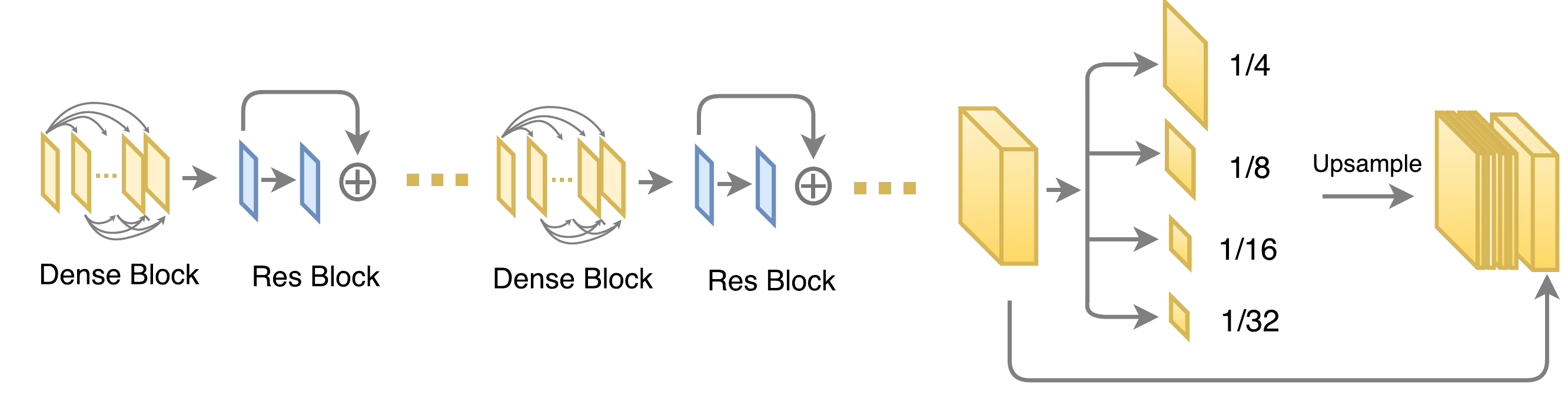}
	\vskip -5pt  \caption{Overview of the dense-residual decoding part.}
	\label{fig:overview_decoder}
\end{figure}

\noindent {\bf{Dense-Residual Decoder.}}
The fused feature representations are then fed into a decoder network that is based on dense-residual decoding blocks. Specifically, the decoder contains five dense-residual blocks, where each dense-residual block contains a dense block, a transition up block and two residual blocks.  Each dense block has the same structure as the dense block described in the encoder.   Each transition up layer  is composed of a 1$\times$1 convolution layer  followed by a bilinear up-sampling  layer.  Each residual blocks contains two 3$\times$3 convolution layer connected by the ReLU function and Batch normalization. Once the feature maps are up-sampled to the original resolution (input resolution, e.g. $256\times256$), these learned features are concatenated with the three input Stokes images. Finally, a multi-level pyramid pooling block is adopted at the end of the decoding part to make sure that features from different scales are embedded in the final result. This is inspired by the use of global context information in classification and segmentation tasks \cite{psp_net,Zhang_2018_CVPR}. Rather than taking very large pooling size to capture more global context information between different objects \cite{psp_net}, more `local' information is leveraged here.  Hence, a four-level pooling operation with  sizes 1/32, 1/16, 1/8 and 1/4 are used. Then, features from all four levels are up-sampled to the original feature size and are concatenated back with the original feature maps before the final estimation.  Figure~\ref{fig:overview_decoder} gives an overview of the proposed dense-residual decoder.

Specifically, the final multi-scale pyramid pooling structure  contains a four scale down-sampling operator followed by a 1x1 convolution layer with one-channel output functioning as depth-wise dimension reduction.  Then, all four scale one-channel feature maps are concatenated with the corresponding input of the multi-scale pooling structure by  up-sampling to the input-feature resolution.  Finally, the concatenated features are fed into a 3$\times$3 convolution layer followed by a Tanh layer. 

\subsection{Multi-scale Discriminator}
To ensure  the synthesized visible faces are indistinguishable from real images while preserving high-frequency details, a learned multi-scale patch-discriminator sub-network is designed to decide if each input image (to the discriminator) is real or fake.  Similar to the structure that was proposed in \cite{pix2pix}, a convolution layer with  batch normalization and Leaky ReLU~\cite{leaky_relu} activation are used as the basis throughout the patch-discriminator part. Basically, the  patch-discriminator consists of the following structure: \\

\emph{CBL($K_2$)-CBL(2$K_2$)-CBL(4$K_2$)-CBL(8$K_2$)} \\

\noindent where, $CBL(K_2)$ is a set of $K_2$-channel convolution layers followed by batch normalization and  Leaky ReLU~\cite{leaky_relu}.  Then,  a multi-scale  pooling module, which pools features at different scales,  is stacked at the end of the discriminator.   The pooled features are then upsampled and concatenated, followed by a 1$\times$1 convolution and a sigmoid function to produce a probability score normalized between 0 and 1.   The proposed discriminator sub-network, $D$, is shown at the bottom of Figure~\ref{fig:overview}.

\subsection{Loss Functions}
It is well-known that the use of Euclidean loss, $L_E$, alone often results in blurry results. Hence, to overcome this and  to discriminate the generated visible face images from their corresponding ground truth, an adversarial loss function is employed. Even though the use of adversarial loss can generate more reasonable results compared to the $L_E$ loss, as shown in \cite{face_ijcb2017}, these results contain undesirable facial artifacts. To address this issue and generate visually pleasing results, perceptual loss is incorporated in our work. The perceptual loss is computed using a pre-trained VGG-16 models as discussed in \cite{perceptual_loss,hang_style,derain_2017_zhang,SR_photorea,chen2017stylebank,he2018deep}. 

Since the ultimate goal of the our proposed synthesis method is to guarantee that human examiners or face verification systems can identify the person given his/her synthesized face images, it is also important to involve the discriminative information into consideration.  Similar to the perceptual loss, we propose an identity-preserving loss that is evaluated on a certain layer of the fine-tuned VGG-Polar model. The VGG-Polar model is fine-tuned using the visible images with their corresponding labels from the newly introduced Polarimetric Visible database.
 
The proposed method contains the following loss functions: the Euclidean $L_{2}$ loss  enforced on the reconstructed visible image, the $L_{E(G)}$ loss enforced on the guidance part,  the adversarial loss to guarantee more sharp and indistinguishable results, the perceptual loss to preserve more photo realistic details and the identity loss to preserve more discriminative information for the outputs. The overall loss function is defined as follows

\begin{equation}
\label{eq:overall_loss}
L_{\text{all}} = L_{2} + L_{2(G)}+ \lambda_A L_{A} + \lambda_PL_{P} +\lambda_IL_{I} ,
\end{equation}
where $L_{2}$ denotes the Euclidean  loss, $L_{2(G)}$ denotes the Euclidean  loss on the guidance sub-network, $L_A$ represents the adversarial loss, $L_P$ indicates the perceptual loss and $L_I$ is the identity loss. Here, $\lambda_A$, $\lambda_P$ and $\lambda_I$ are the corresponding weights. 

The $L_2$ and the adversarial losses are defined as follows:
\begin{equation}
\label{eq:trans_loss_euc}
 L_{2}, L_{2(G)}= \sum_{w,h}^{}\|\phi_{G}({S_0,S_1,S_2})^{w,h}-Y_t^{w,h}\|_2,	\\
 \end{equation} 
 \begin{equation}
 \label{eq:trans_loss_adv}
   L_A= -\log(\phi_D(\phi_G({S_0,S_1,S_2})),
 \end{equation}
 where $S_0$, $S_1$ and $S_2$ are the three different input Stokes images, $Y_t$ is the ground truth visible image, $W\times H$ is the dimension of the input image, $\phi_{G}$ is the multi-stream feature-fusion generator sub-network $G$ and $\phi_D$ is the multi-scale discriminator sub-network $D$.

As the perceptual loss and the identity losses are evaluated on a certain layer of the given CNN model, both can be defined as follows:  
\begin{equation}
\label{eq:dehaze_loss_perc}
 L_{P,I}= \sum_{c_i,w_i,h_i}^{} \|V(\phi_G({S_0,S_1,S_2}))^{c_i,w_i,h_i}-V
 (Y_t)^{c_i,w_i,h_i}\|_2,
\end{equation}
 where $Y_t$ is the ground truth visible image, $\phi_{E}$ is the proposed generator, $V$ represents a non-linear CNN transformation and $C_i,W_i,H_i$ are the dimensions of a certain high level layer $V$, which differs for perceptual and identity losses.

 \section{Polarimetric Thermal Face Dataset}
 \label{sec:database}
 A polarimetric thermal face database of 111 subjects is used for this study, which expanded on the previously released database of 60 subjects (described in detail in Hu \emph{et al.}, 2016~\cite{facedata_cvprw17}).  The database used in this study therefore consisted of the 60-subject database collected at the U.S. Army Research Laboratory (ARL) in 2014-2015 (referred to as Volume 1 hereinafter), and a 51-subject database collected at a Department of Homeland Security test facility (referred to as Volume 2 hereinafter).  While the participants of the Volume 1 collect consisted exclusively of ARL employees, the participants of the Volume 2 collect were recruited from the local community in Maryland, resulting in more demographic diversity.  Note that this  extended databased is available upon request.

 \subsection{Sensors}
 \label{ssec:sensors}
 The sensors employed to collect Volume 1 and Volume 2 were the same, consisting of a polarimetric LWIR imager and visible cameras.  The LWIR polarimetric was developed by Polaris Sensor Technologies, and is based on a division-of-time spinning achromatic retarder (SAR) design which incorporated a spinning phase-retarder in conjunction with a linear wire-grid polarizer.  This system has a spectral response range of 7.5-11 $\mu m$, and employed a Stirling cooler with a mercury telluride focal plane array ($640 \times 480$ pixel array format).  Data was recorded at 60 frames per second, using a lens with a field of view (FOV) of $10.6\degree \times 7.9\degree$.   Four Basler Scout GigE cameras with different lens (ranging from $5\degree$ to $53\degree$) were used for Volume 1, consisting of two grayscale cameras (model \# scA640-70gm; $659 \times 494$ pixel FPA) and two color cameras (model \# scA640-70gc; $658 \times 492$ pixel FPA) to generate visible facial imagery at different resolutions.  For Volume 2, a single Basler Scout color camera with a zoom lens was used, adjusted to produce the same facial resolution as the polarimeter.
  \begin{figure}[htp!]
  \centering
  \includegraphics[width=0.5\textwidth]{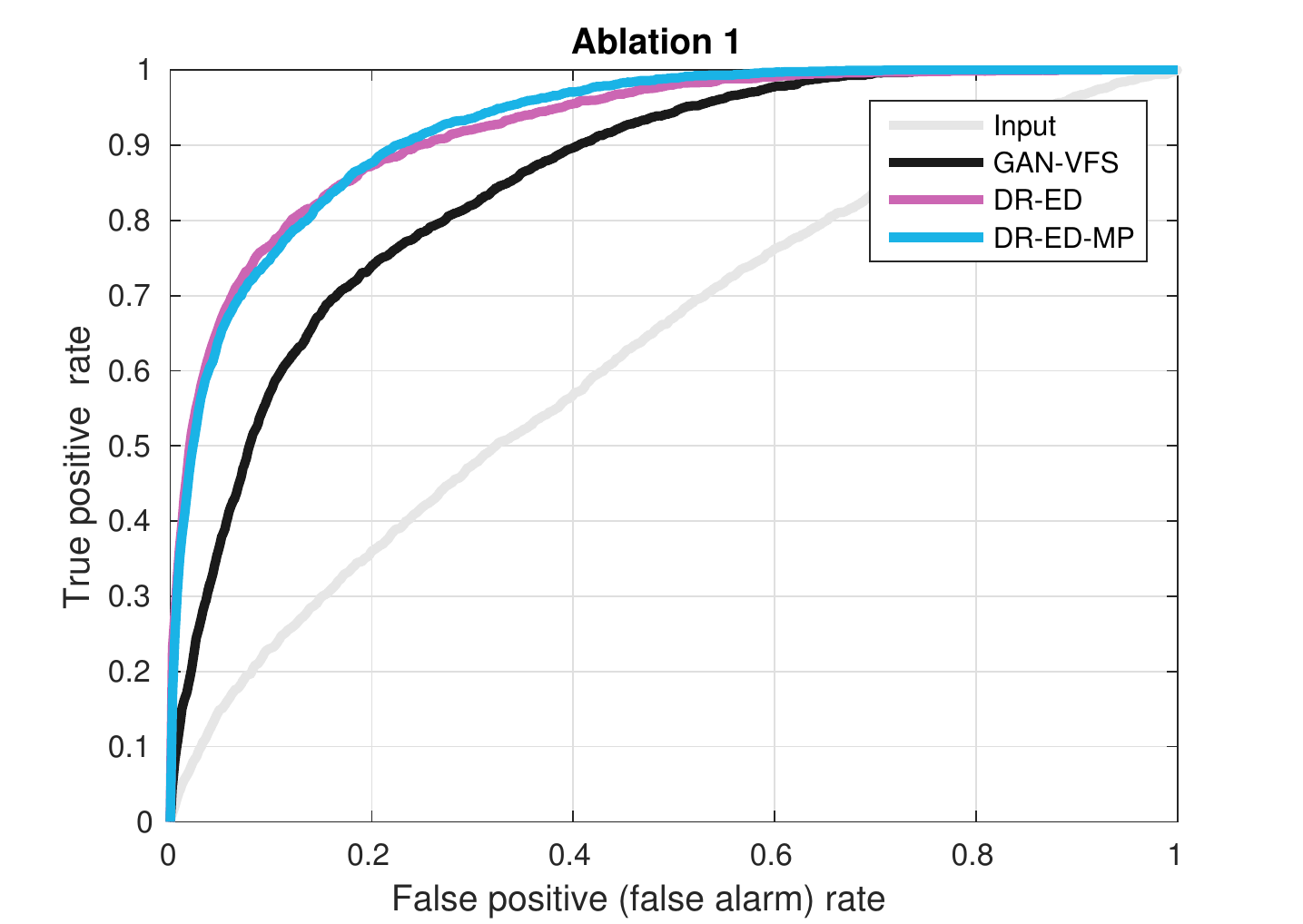}
   \vskip -6pt  \caption{The ROC curves corresponding to \textbf{Ablation 1}. }
  \label{fig:ab1}
  \end{figure}
  \begin{figure*}[t]
  	\centering
  		\begin{minipage}{.19\textwidth}
  			\centering
  			\caption*{PSNR:11.55;\\ SSIM: 0.46} \vskip-10pt
  			\includegraphics[width=1\textwidth]{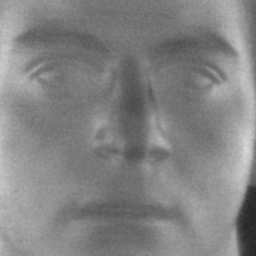}
  			\captionsetup{labelformat=empty}
  			\captionsetup{justification=centering}
  	\end{minipage}
  	\begin{minipage}{.19\textwidth}
  		\centering
  			\caption*{PSNR:19.42;\\ SSIM: 0.75} \vskip-10pt
  			\includegraphics[width=1\textwidth]{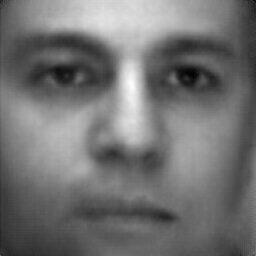}
  		\captionsetup{labelformat=empty}
  		\captionsetup{justification=centering}
  	\end{minipage}
  	\begin{minipage}{.19\textwidth}
  		\centering
  			\caption*{PSNR:19.82;\\ SSIM: 0.78} \vskip-10pt
  			\includegraphics[width=1\textwidth]{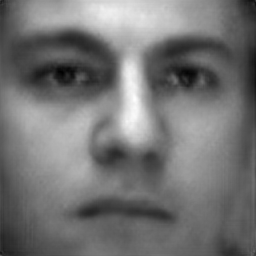}
  		\captionsetup{labelformat=empty}
  		\captionsetup{justification=centering}
  	\end{minipage}
  	\begin{minipage}{.19\textwidth}
  		\centering
  			\caption*{PSNR:\textbf{21.32};\\ SSIM: \textbf{0.80}} \vskip-10pt
  			\includegraphics[width=1\textwidth]{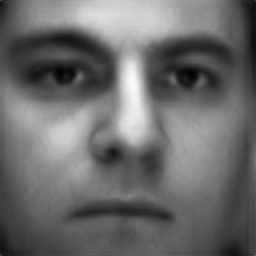}
  		\captionsetup{labelformat=empty}
  		\captionsetup{justification=centering}
  	\end{minipage}
  	\begin{minipage}{.19\textwidth}
  		\centering
  		\caption*{PSNR:Inf;\\ SSIM: 1} \vskip-10pt
  			\includegraphics[width=1\textwidth]{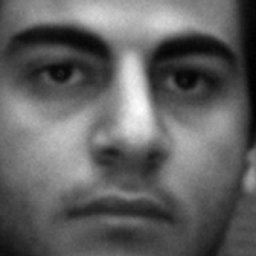}
  		\captionsetup{labelformat=empty}
  		\captionsetup{justification=centering}
  	\end{minipage}		\vskip+6pt
  		\begin{minipage}{.19\textwidth}
  			\centering
  			\includegraphics[width=1\textwidth]{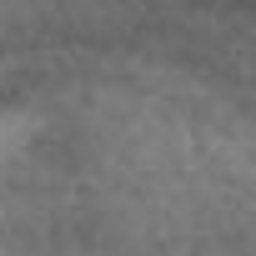}
  			\captionsetup{labelformat=empty}
  			\captionsetup{justification=centering}
  			\caption*{I-Polar}
  	\end{minipage}
  	\begin{minipage}{.19\textwidth}
  		\centering
  			\includegraphics[width=1\textwidth]{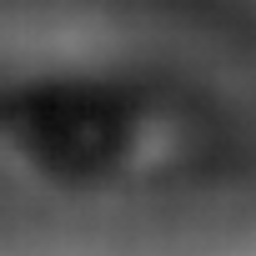}
  		\captionsetup{labelformat=empty}
  		\captionsetup{justification=centering}
  			\caption*{GAN-VFS \cite{face_ijcb2017}}
  	\end{minipage}
  	\begin{minipage}{.19\textwidth}
  		\centering
  			\includegraphics[width=1\textwidth]{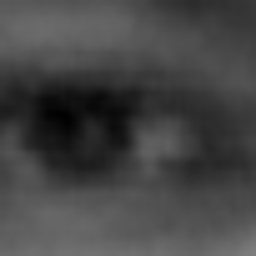}
  		\captionsetup{labelformat=empty}
  		\captionsetup{justification=centering}
  		\caption*{DR-ED}
  	\end{minipage}
  	\begin{minipage}{.19\textwidth}
  		\centering
  			\includegraphics[width=1\textwidth]{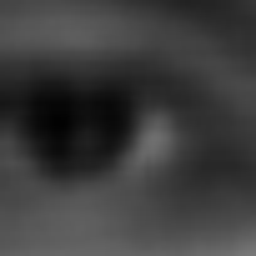}
  		\captionsetup{labelformat=empty}
  		\captionsetup{justification=centering}
  		\caption*{DR-ED-MP}
  	\end{minipage}
  	\begin{minipage}{.19\textwidth}
  		\centering
  			\includegraphics[width=1\textwidth]{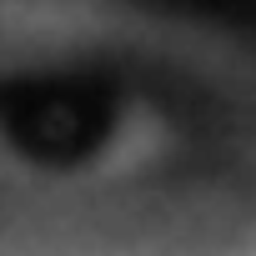}
  		\captionsetup{labelformat=empty}
  		\captionsetup{justification=centering}
  		\caption*{Target}
  	\end{minipage}
  	\vskip-6pt
  	\caption{Sample results of for the \textbf{Ablation 1}. It can be observed that the dense-resisual encoder-decoder structure is able to generate better visible results and the introduced multi-level pooling module is able to preserve better structure information.   Detail discussions can be found in Sec~\ref{ssec: ablaion}.}  \label{fig:ablation1}
  \end{figure*}

 \subsection{Dataset}
 \label{ssec:protocol}
 The dataset protocols for Volume 1 and Volume 2 were approved by the respective Institutional Review Boards (IRBs) where each collection occurred.   The Volume 1 collection involved two experimental conditions: range and expressions.  Acquisitions were made at distances of 2.5 m, 5 m, and 7.5 m.  At each range, a 10 second video sequence was first collected of the subject with a neural expression, and then a 10 second ``expressions'' sequence was collected as the subject counted out loud numerically from one upwards, which induced a continuous range of motions of the mouth and, to a lesser extent the eyes.  In the experimental setup for Volume 1, a floor lamp was placed 1 m in front of the subject at each range to provide additional illumination.
   \begin{table}[h]
   \centering
   \caption{The average PSNR (dB), SSIM, EER and AUC results corresponding to different methods for \textbf{Ablation 1}.}
   \label{ta:ab1_result_psnr}
   \resizebox{.48\textwidth}{!}{%
   \begin{tabular}{|c|c|c|c|c|}
   \hline
    & I-Polar & GAN-VFS \cite{face_ijcb2017} & DR-ED & DR-ED-MP \\ \hline
   PSNR (dB) & 11.74 & 18.07 & 18.28 & \textbf{18.80}\\ \hline
   SSIM & 0.4625 & 0.7047 & 0.7128 & \textbf{0.7194} \\ \hline
      EER & 41.51\% & 22.45\% & 16.51\% & \textbf{15.67\%}\\ \hline
   AUC &  62.93\% & 86.10\% & 91.67\% & \textbf{92.55\%} \\ \hline
   \end{tabular}%
   }
   \end{table}
   

 The data collection setup used for Volume 2 matched that of Volume 1.  However, no floor lamp was employed in the Volume 2 collect, as the DHS test facility had sufficient illumination.  Furthermore, Volume 2 data was collected at a single range of 2.5 m, due to time limitations since the polarimetric face acquisition was part of a broader collection.
 \subsection{Preprocessing}
 \label{ssec:preprocessing}
 The raw polarimetric thermal imagery underwent several preprocessing steps.  First, a two-point non-uniformity correction (NUC) was applied on the raw data using software provided by Polaris Sensor Technologies and calibration data collected with a Mikron blackbody prior to each session.  Images were sampled/extracted from the polarimetric thermal sequences.  Bad pixels in the extracted images were identified, and those pixel intensities corrected via a median filter.  To crop and align the facial imagery, three fiducial points (centers of the eyes, base of the nose) were first manually annotated, and an affine transform was used to normalize each face to canonical coordinates.  Facial imagery was finally cropped to $m \times n$ pixels, and saved as 16-bit PNG files.  The visible imagery required neither non-uniformity correction nor bad pixel correction. The same steps were used to crop and align the visible images, which were then saved as 16-bit grayscale PNG files.

\subsection{Experimental Protocols}
\label{ssec:databases_set}
Even though there exist several conventional thermal-visible pair databases \cite{face_thermal_dataset1,face_thermal_dataset2}, they lack the availability of the corresponding polarization state information such as $S_1$ and $S_2$. Hence, an extended database, which  contains polarimetric ($S_0$, $S_1$, $S_2$) and visible image pairs from 111 subjects is used for evaluation in this paper.     Following the protocol defined in  \cite{face_btas2016,face_ijcb2017}, sample pairs corresponding to range 1 (baseline and expression) are used for comparisons.   In particular, two different protocols are defined in this paper for further research.
To be consistent with previous methods \cite{face_btas2016,face_ijcb2017},  the first protocol is defined  as follows: \\

\noindent {\bf{Protocol 1:}} The protocol 1 is evaluated on Volume 1, which contains 60 subjects,  30 subjects from Volume 1  with eight samples for each subject (in total 240 sample pairs) are used as training samples, denoted as \emph{Train1}. Similarly, the remaining 30 subjects with eight samples for each subject (in total 240 sample pairs) are used as testing samples, denoted as \emph{Protocol1}. All the training and testing samples are randomly chosen from the overall 60 subjects. Results are evaluated on five random splits.  In Protocol 1,  each split contains around 28800 pairs of templates on average (1080 positive and 27720 negative).  \\

\noindent {\bf{Protocol 2:}} Different from Protocol 1, the newly introduced and extended dataset with 111 subjects is used for training and testing, where  85 subjects with eight samples for each subject are randomly chosen  as training samples (in total 680 sample pairs), denoted as \emph{Train2} and the other 26 subjects are used as testing (in total 208 sample pairs), denoted as \emph{Protocol2}.  As before, results are evaluated on five random splits. In Protocol 2,  each split on average contains around 21632 pairs of templates (936 positive and 20696 negative). 

These protocols and splits will be made publicly available to the research community.

\section{Experimental Results}
\label{sec:experiment}
In this section, we demonstrate the effectiveness of the
proposed approach by conducting various experiments on 
the two defined protocols for the new polarimetric thermal dataset as described above.  Once the visible images are synthesized using the proposed method, deep features can be extracted from these images using any one of many pre-trained CNNs such as VGG-face \cite{vggface}, Light-CNN \cite{light_CNN}, or GoogleNet \cite{google_face}.  In this paper, we extract the features from the second last fully connected layer of the VGG-face network \cite{vggface}.  Finally, the cosine distance is used to calculate the scores.  Results are compared with four state-of-the-art methods: Ben {\emph{et al.}} \cite{face_btas2016}, GAN-VFS  \cite{face_ijcb2017}, Pix2pix \cite{pix2pix} and Pix2pix with BEGAN~\cite{pix2pix,began}. In addition, three ablation studies are conducted to demonstrate the effectiveness of  different modules of the proposed method.  Quality of the synthesized images is evaluated using Peak Signal-to-Noise Ratio (PSNR) and Structural SIMilarity (SSIM) index \cite{ssim}.  The face verification performance is evaluated using the receiver operating characteristic (ROC) curve, Area Under the Curve (AUC) and Equal Error Rate (EER) measures.

\begin{figure*}[t]
	\centering
		\begin{minipage}{.15\textwidth}
			\centering
			\caption*{PSNR: 12.88 \\ SSIM: 0.5911}\vskip-10pt
			\includegraphics[width=1\textwidth]{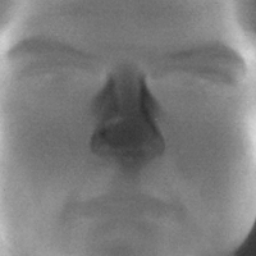}
			\captionsetup{labelformat=empty}
			\captionsetup{justification=centering}
		\caption*{I-Polar}
	\end{minipage}
	\begin{minipage}{.15\textwidth}
		\centering
			\caption*{PSNR: 9.08 \\ SSIM: 0.5623}\vskip-10pt
			\includegraphics[width=1\textwidth]{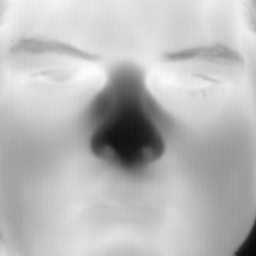}
		\captionsetup{labelformat=empty}
		\captionsetup{justification=centering}
		\caption*{I-$S_0$}
	\end{minipage}
	\begin{minipage}{.15\textwidth}
		\centering
			\caption*{PSNR:10.64 \\ SSIM: 0.4863}\vskip-10pt
			\includegraphics[width=1\textwidth]{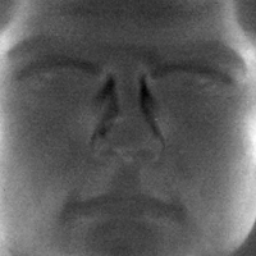}
		\captionsetup{labelformat=empty}
		\captionsetup{justification=centering}
		\caption*{I-$S_1$}
	\end{minipage}
	\begin{minipage}{.15\textwidth}
		\centering
			\caption*{PSNR: 13.08 \\ SSIM: 0.5508}\vskip-10pt
			\includegraphics[width=1\textwidth]{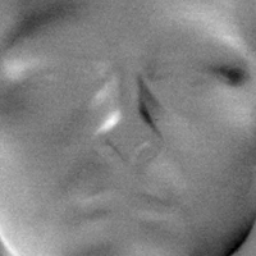}
		\captionsetup{labelformat=empty}
		\captionsetup{justification=centering}
		\caption*{I-$S_2$}
	\end{minipage}
	\begin{minipage}{.15\textwidth}
		\centering
			\caption*{PSNR:Inf \\ SSIM: 1.0000}
\vskip-10pt
			\includegraphics[width=1\textwidth]{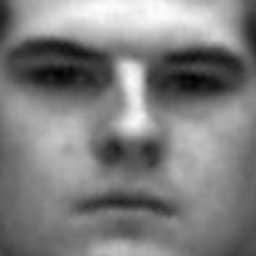}
		\captionsetup{labelformat=empty}
		\captionsetup{justification=centering}
		\caption*{Target}
	\end{minipage}\\	\vskip-8pt
	\begin{minipage}{.15\textwidth}
		\centering
		\caption*{PSNR:18.83\\ SSIM: 0.8004}\vskip-10pt
			\includegraphics[width=1\textwidth]{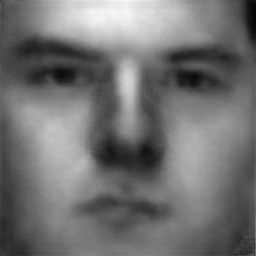}
		\captionsetup{labelformat=empty}
		\captionsetup{justification=centering}
		\caption*{S-$S_0$}
	\end{minipage}
	\begin{minipage}{.15\textwidth}
		\centering
			\caption*{PSNR:19.34\\ SSIM: 0.7904}\vskip-10pt
			\includegraphics[width=1\textwidth]{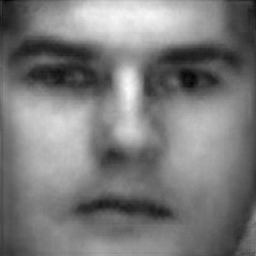}
		\captionsetup{labelformat=empty}
		\captionsetup{justification=centering}
		\caption*{S-$S_1$}
	\end{minipage}
	\begin{minipage}{.15\textwidth}
		\centering
			\caption*{PSNR:20.43\\ SSIM: 0.8143}\vskip-10pt
			\includegraphics[width=1\textwidth]{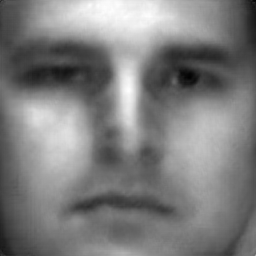}
		\captionsetup{labelformat=empty}
		\captionsetup{justification=centering}
		\caption*{S-$S_2$}
	\end{minipage}
	\begin{minipage}{.15\textwidth}
		\centering
			\caption*{PSNR:20.56\\ SSIM: 0.8367}\vskip-10pt
			\includegraphics[width=1\textwidth]{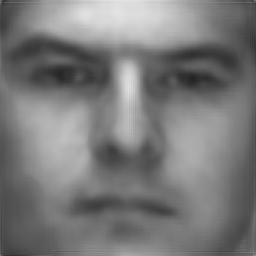}
		\captionsetup{labelformat=empty}
		\captionsetup{justification=centering}
		\caption*{S-Polar-IF}
	\end{minipage}
	\begin{minipage}{.15\textwidth}
		\centering
			\caption*{PSNR:20.85\\ SSIM: 0.8512}\vskip-10pt
			\includegraphics[width=1\textwidth]{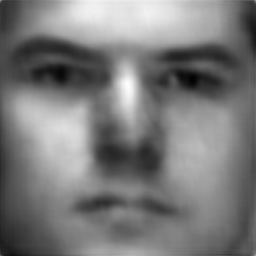}
		\captionsetup{labelformat=empty}
		\captionsetup{justification=centering}
		\caption*{M-Polar-OF}
	\end{minipage}
	\begin{minipage}{.15\textwidth}
		\centering
			\caption*{PSNR:\textbf{21.65}\\ SSIM: \textbf{0.8622}}\vskip-10pt
			\includegraphics[width=1\textwidth]{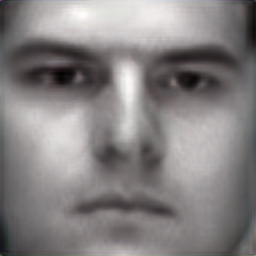}
		\captionsetup{labelformat=empty}
		\captionsetup{justification=centering}
		\caption*{Proposed}
	\end{minipage}
	\vskip-15pt
	\caption{Sample results for \textbf{Ablation 2}. It can be observed that the proposed multi-stream feature-level fusion GAN  is able to generate better results compared to input-level (S-Polar-IF), output-level fusion (M-Polar-OF) and also simply levering single Stokes modality. Detailed discussions can be found in Sec~\ref{ssec: ablaion}.}  \label{fig:ablation2}
\end{figure*}
\begin{table}[htp!]
\centering
\caption{The average PSNR (dB), SSIM, EER and AUC results corresponding to different methods for \textbf{Ablation 2}.}
\label{tab:ablation2}
\resizebox{.5\textwidth}{!}{%
\begin{tabular}{|c|c|c|c|c|c|c|c|}
\hline
  & I-Polar & S-$S_0$ & S-$S_1$ & S-$S_2$ & S-Polar-IF & M-Polar-OF & Proposed  \\ \hline
PSNR (dB) & 11.74 & 17.34 & 17.03 & 17.17 & {18.80} & {18.87} & \textbf{19.55}\\ \hline
SSIM & 0.4625 & 0.6905 & 0.6852 & 0.6794 & {0.7194} & {0.7225}& \textbf{0.7433} \\ \hline
EER & 41.51\% & 23.18\% & 21.61\% & 21.56\% & 15.67\% & 15.90\% & \textbf{11.78}\%\\ \hline
AUC & 62.93\% & 85.74\% & 86.64\% & 87.30\% & 92.55\% & 92.69\% & \textbf{96.03}\% \\ \hline
\end{tabular}%
}
\end{table}

\subsection{Implementation}
The entire network is trained on a Nvidia Titan-X  GPU. We choose $\lambda_A=0.005$ for the adversarial loss,  $\lambda_P=0.8$  for the perceptual loss and  $\lambda_I=0.1$ for the identity loss. During training, we use ADAM \cite{adam_opt} as the optimization algorithm with learning rate  of $8\times 10^{-4}$ and batch size of  1 image.  All the pre-processed training samples are resized to $256\times 256$. The perceptual loss is evaluated on relu 1-1 and relu 2-1 layers in the pre-trained VGG \cite{vggface} model.  The identity loss is evaluated on the relu2-2 layer of the fine-tuned VGG-Polar model.

\subsection{Ablation Study} 
\label{ssec: ablaion}
In order to better demonstrate the effectiveness of the proposed feature-level fusion, the improvements obtained by different modules and the importance of different loss functions in  the proposed network, three ablation studies are presented in this section. All the experiments in the first two ablation studies are optimized with the same loss function discussed in Eq~(2).\\

\begin{figure*}[t]
	\centering
		\begin{minipage}{.13\textwidth}
			\centering
			\caption*{PSNR: 10.80 \\ SSIM: 0.4356}\vskip-10pt
			\includegraphics[width=1\textwidth]{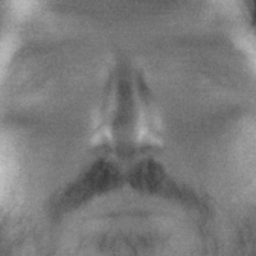}
			\captionsetup{labelformat=empty}
			\captionsetup{justification=centering}
			\caption*{Input}
	\end{minipage}
	\begin{minipage}{.13\textwidth}
		\centering
			\caption*{PSNR:17.11 \\ SSIM: 0.6953}\vskip-10pt
			\includegraphics[width=1\textwidth]{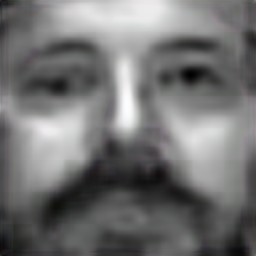}
		\captionsetup{labelformat=empty}
		\captionsetup{justification=centering}
			\caption*{L2}
	\end{minipage}
	\begin{minipage}{.13\textwidth}
		\centering
			\caption*{PSNR:16.99 \\ SSIM: 0.7032}\vskip-10pt
			\includegraphics[width=1\textwidth]{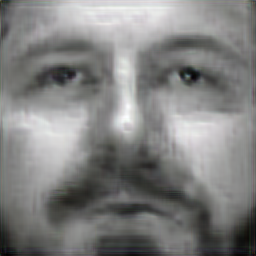}
		\captionsetup{labelformat=empty}
		\captionsetup{justification=centering}
			\caption*{L2-GAN}
	\end{minipage}
	\begin{minipage}{.13\textwidth}
		\centering
			\caption*{PSNR:17.88 \\ SSIM: 0.7303}\vskip-10pt
			\includegraphics[width=1\textwidth]{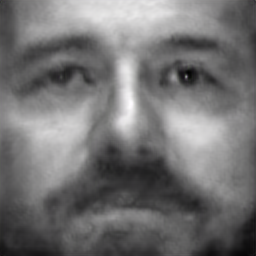}
		\captionsetup{labelformat=empty}
		\captionsetup{justification=centering}
			\caption*{L2-GAN-P}
	\end{minipage}
	\begin{minipage}{.13\textwidth}
		\centering
			\caption*{PSNR: \textbf{18.27} \\ SSIM: \textbf{0.7503}}\vskip-10pt
			\includegraphics[width=1\textwidth]{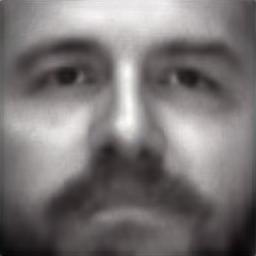}
		\captionsetup{labelformat=empty}
		\captionsetup{justification=centering}
			\caption*{Our}
	\end{minipage}
	\begin{minipage}{.13\textwidth}
		\centering
			\caption*{PSNR:Inf \\ SSIM: 1.0000}\vskip-10pt
			\includegraphics[width=1\textwidth]{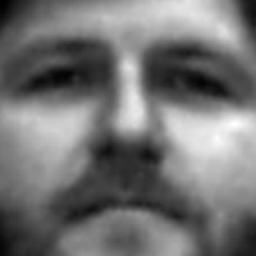}
		\captionsetup{labelformat=empty}
		\captionsetup{justification=centering}
			\caption*{Target}
	\end{minipage}
\vskip-5pt
	\caption{Sample results on different loss functions for \textbf{Ablation 3}. } 
\end{figure*}

\noindent {\bf{Ablation 1}}\\
In the first ablation study, we demonstrate the effectiveness of different modules (eg. densely connected encoder-decoder structure) in our method by conducting the following experiments. All the experimental results are evaluated using \textbf{Protocol 1} based on the polrimetric images as input:   
\begin{enumerate}[nolistsep]
\item[(a)] \textbf{GAN-VFS}: The GAN network proposed in \cite{face_ijcb2017} with polarimetric images as inputs. 
\item[(b)] \textbf{DR-ED}: A single stream dense-resisual encoder-decoder structure. \footnote{Basically, this network is composed of  one stream of the encoder part followed by the same decoder without multi-level pooling.}
\item[(c)] \textbf{DR-ED-MP}: A single stream dense-resisual encoder-decoder structure with multi-level pooling.

\end{enumerate}
\begin{figure}[htp!]
\centering
\includegraphics[width=0.5\textwidth]{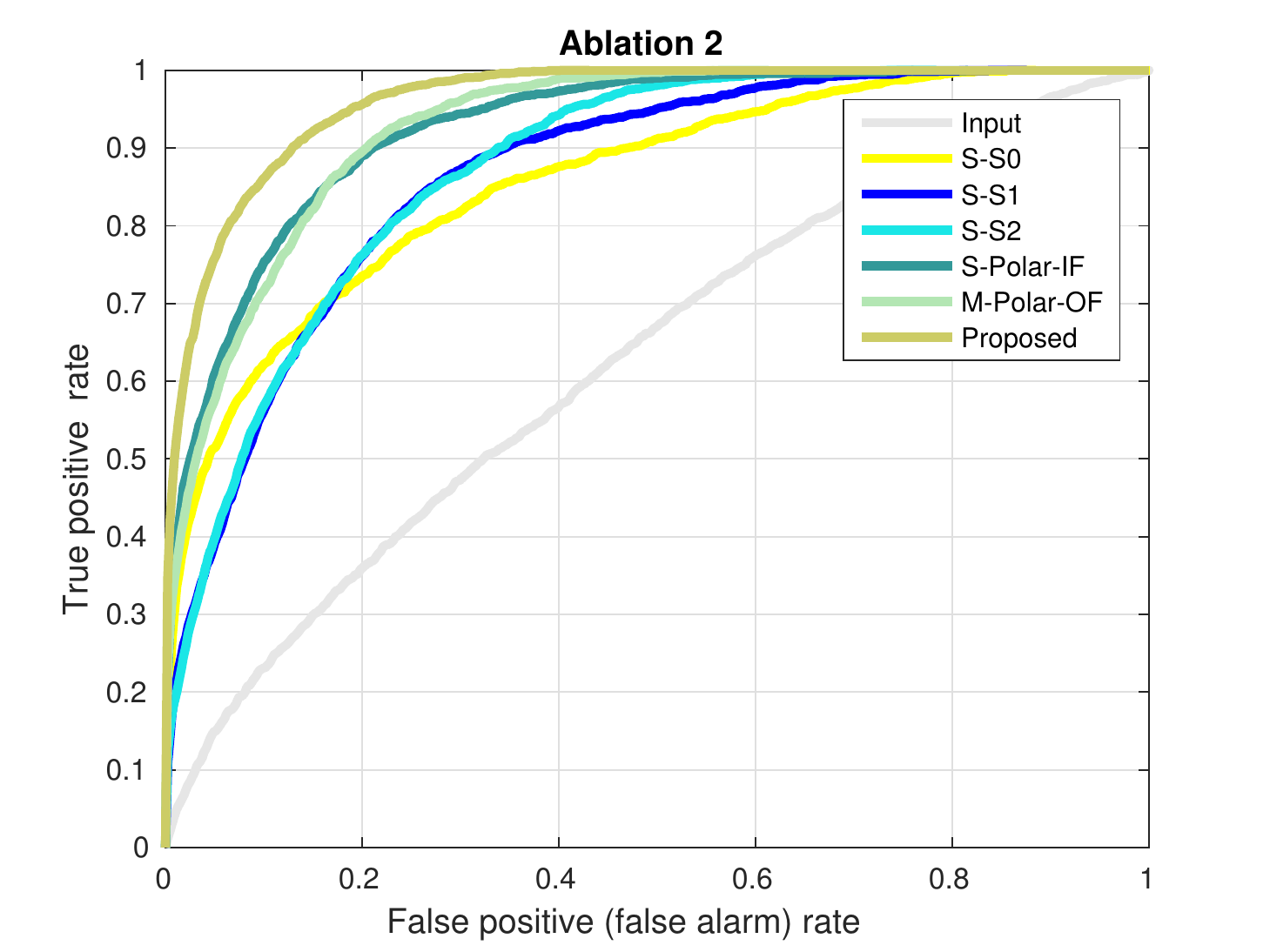}
 \vskip -6pt  \caption{The ROC curves corresponding to \textbf{Ablation 2}. }
\label{fig:ab2}
\end{figure}
\begin{figure}[htp!]
	\centering
	\includegraphics[width=0.5\textwidth]{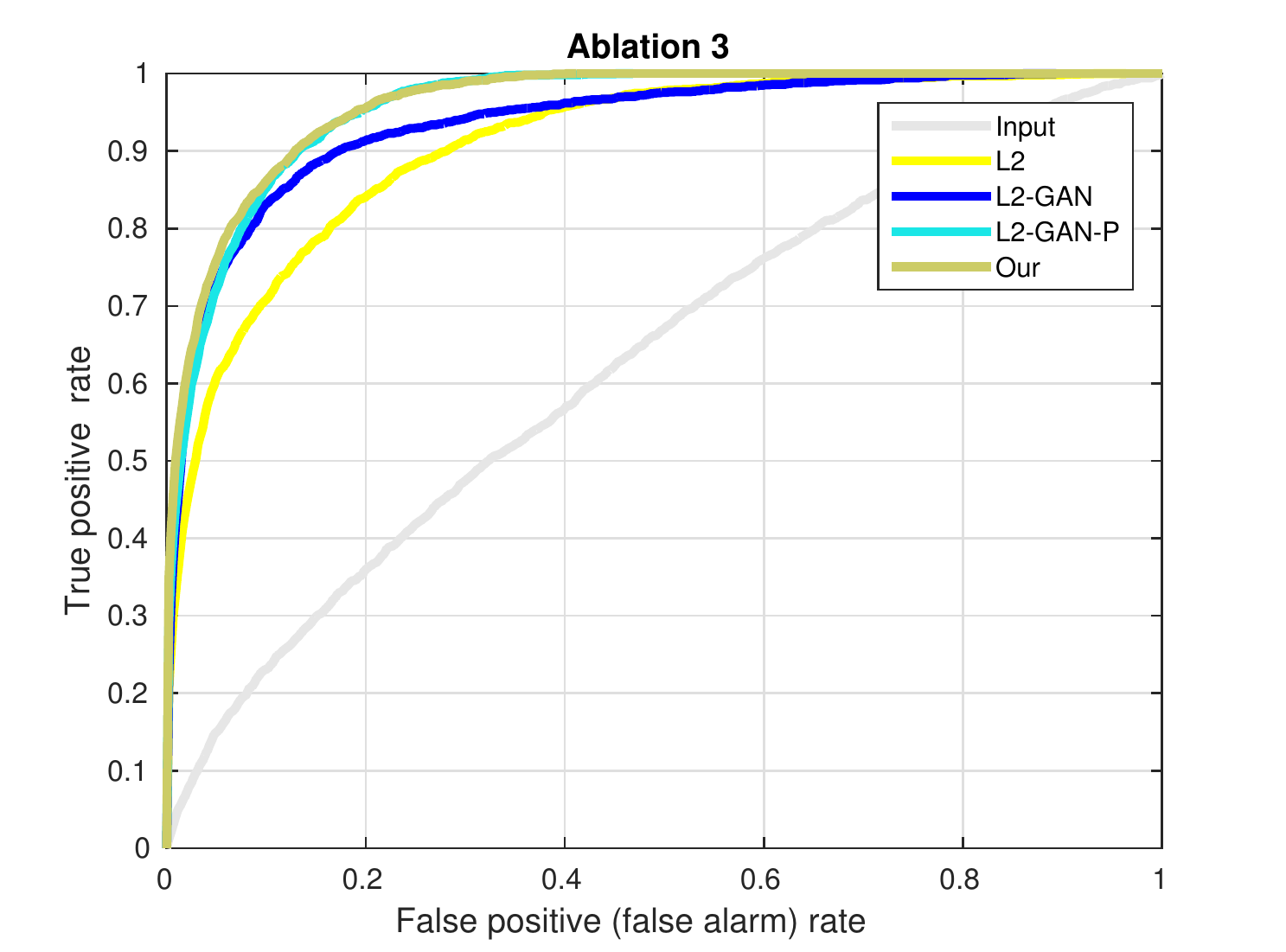}
	\vskip -6pt  \caption{The ROC curves corresponding to \textbf{Ablation 3}. }
	\label{fig:ab32}
\end{figure}

\begin{table}[h]
	\centering
	\caption{The average PSNR (dB), SSIM, EER and AUC results corresponding to different methods for \textbf{Ablation 3}.}
	\label{ta:ab3_result_psnr}
	\resizebox{.48\textwidth}{!}{%
		\begin{tabular}{|c|c|c|c|c|c|}
			\hline
			& I-Polar & L2 & L2-GAN & L2-GAN-P & Our \\ \hline
			PSNR (dB) & 11.74 & 17.57 & 17.33 & {18.99} & \textbf{19.55}\\ \hline
			SSIM & 0.4625 & 0.7088 & 0.7115 & {0.7352} & \textbf{0.7433} \\ \hline
			EER & 41.51\% & 18.07\% & 13.23\% & {11.79\%} & \textbf{11.78\%}\\ \hline
			AUC &  62.93\% & 90.89\% & 93.64\% & {95.64\%} & \textbf{96.03\%}\\ \hline
		\end{tabular}%
	}
\end{table}

\begin{figure}[h]
\centering
\includegraphics[width=0.5\textwidth]{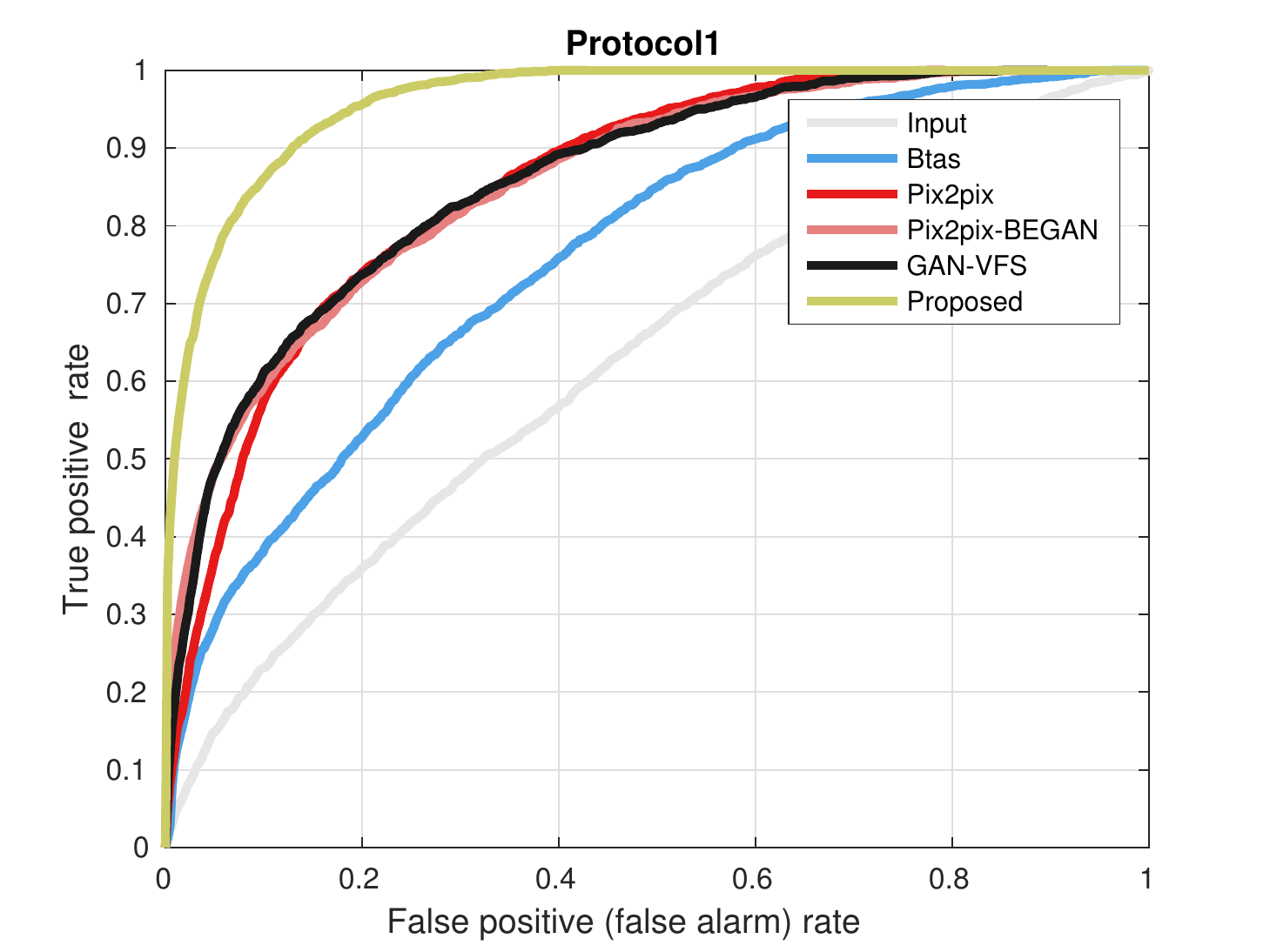}
 \vskip -6pt  \caption{The ROC curves corresponding to \emph{Protocol1}. }
\label{fig:ab3}
\end{figure}

One synthesis example corresponding to \textbf{Ablation 1} is shown in Figure~\ref{fig:ablation1}. It can be observed from this figure (comparing second column with third column) that the overall performance improves after leveraging the newly introduced dense-residual encoder-decoder (DR-ED) structure.
This can be  clearly observed from the left part of the reconstructed mouth. This essentially demonstrates the effectiveness of the proposed dense-residual encoder-decoder structure.  Though  the DR-ED is able to reconstruct better visible face, from the close-up of the left eye shown in the second row in Figure\ref{fig:ablation1} we observe that some structure information is missing. The multi-level pooling module  at the end of the encoder-decoder structure overcomes this issue and preserves the the overall eye structure.  Quantitative results evaluated based on PSNR and SSIM \cite{ssim}, as shown in Table~\ref{ta:ab1_result_psnr}, also show similar results.

\begin{figure*}[t]
	\centering
		\begin{minipage}{.13\textwidth}
			\centering
			\caption*{PSNR: 10.80 \\ SSIM: 0.4356}\vskip-10pt
			\includegraphics[width=1\textwidth]{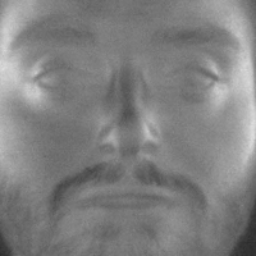}
			\captionsetup{labelformat=empty}
			\captionsetup{justification=centering}
	\end{minipage}
	\begin{minipage}{.13\textwidth}
		\centering
			\caption*{PSNR:15.77 \\ SSIM: 0.6423}\vskip-10pt
			\includegraphics[width=1\textwidth]{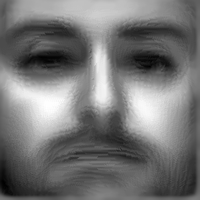}
		\captionsetup{labelformat=empty}
		\captionsetup{justification=centering}
	\end{minipage}
	\begin{minipage}{.13\textwidth}
		\centering
			\caption*{PSNR:16.65 \\ SSIM: 0.6681}\vskip-10pt
			\includegraphics[width=1\textwidth]{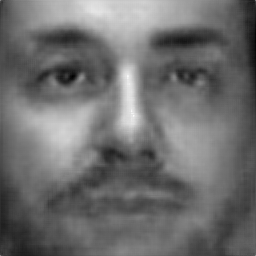}
		\captionsetup{labelformat=empty}
		\captionsetup{justification=centering}
	\end{minipage}
	\begin{minipage}{.13\textwidth}
		\centering
			\caption*{PSNR:17.05 \\ SSIM: 0.6603}\vskip-10pt
			\includegraphics[width=1\textwidth]{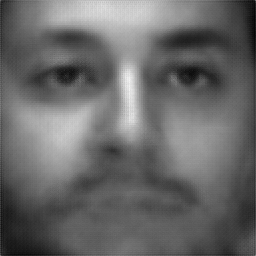}
		\captionsetup{labelformat=empty}
		\captionsetup{justification=centering}
	\end{minipage}
	\begin{minipage}{.13\textwidth}
		\centering
			\caption*{PSNR:19.03 \\ SSIM: 0.7203}\vskip-10pt
			\includegraphics[width=1\textwidth]{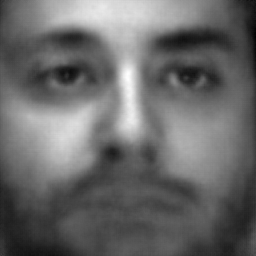}
		\captionsetup{labelformat=empty}
		\captionsetup{justification=centering}
	\end{minipage}
	\begin{minipage}{.13\textwidth}
		\centering
			\caption*{PSNR:\textbf{19.52} \\ SSIM: \textbf{0.7380}}\vskip-10pt
			\includegraphics[width=1\textwidth]{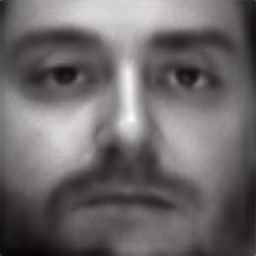}
		\captionsetup{labelformat=empty}
		\captionsetup{justification=centering}
	\end{minipage}
	\begin{minipage}{.13\textwidth}
		\centering
			\caption*{PSNR:Inf \\ SSIM: 1.0000}
\vskip-10pt
			\includegraphics[width=1\textwidth]{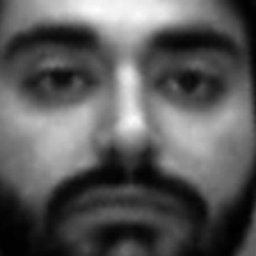}
		\captionsetup{labelformat=empty}
		\captionsetup{justification=centering}
	\end{minipage}\\	\vskip+5pt
		\begin{minipage}{.13\textwidth}
			\centering
			\caption*{PSNR: 10.23 \\ SSIM: 0.4152}\vskip-10pt
			\includegraphics[width=1\textwidth]{result_vol1//189_input.png}
			\captionsetup{labelformat=empty}
			\captionsetup{justification=centering}
		\caption*{I-Polar  \\ \quad}
	\end{minipage}
		\begin{minipage}{.13\textwidth}
			\centering
			\caption*{PSNR: 16.04 \\ SSIM: 0.6122}\vskip-10pt
			\includegraphics[width=1\textwidth]{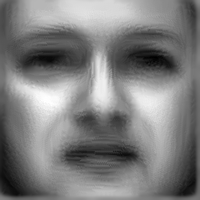}
			\captionsetup{labelformat=empty}
			\captionsetup{justification=centering}
		\caption*{Btas-2016 \cite{face_btas2016}  \\ \quad}
	\end{minipage}
	\begin{minipage}{.13\textwidth}
		\centering
			\caption*{PSNR:17.04 \\ SSIM: 0.6604}\vskip-10pt
			\includegraphics[width=1\textwidth]{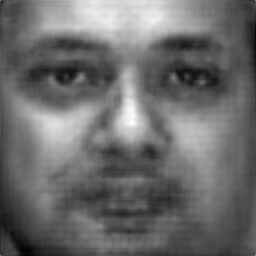}
		\captionsetup{labelformat=empty}
		\captionsetup{justification=centering}
		\caption*{Pix2pix \cite{pix2pix} \\ \quad}
	\end{minipage}
	\begin{minipage}{.13\textwidth}
		\centering
			\caption*{PSNR:16.84 \\ SSIM: 0.6449}\vskip-10pt
			\includegraphics[width=1\textwidth]{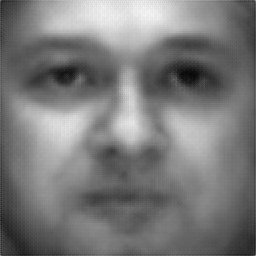}
		\captionsetup{labelformat=empty}
		\captionsetup{justification=centering}
		\caption*{Pix2pix-BEGAN  \\ \quad\cite{pix2pix,began} }
	\end{minipage}
	\begin{minipage}{.13\textwidth}
		\centering
			\caption*{PSNR:17.34 \\ SSIM: 0.6722}\vskip-10pt
			\includegraphics[width=1\textwidth]{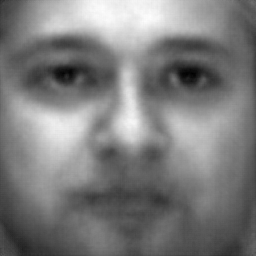}
		\captionsetup{labelformat=empty}
		\captionsetup{justification=centering}
		\caption*{GAN-VFS  \\ \quad\cite{face_ijcb2017}}
	\end{minipage}
	\begin{minipage}{.13\textwidth}
		\centering
			\caption*{PSNR:\textbf{18.05} \\ SSIM: \textbf{0.7134}}\vskip-10pt
			\includegraphics[width=1\textwidth]{result_vol1//189_our.png}
		\captionsetup{labelformat=empty}
		\captionsetup{justification=centering}
		\caption*{Proposed  \\ \quad}
	\end{minipage}
	\begin{minipage}{.13\textwidth}
		\centering
			\caption*{PSNR:Inf \\ SSIM: 1.0000}
\vskip-10pt
			\includegraphics[width=1\textwidth]{result_vol1//189_target.png}
		\captionsetup{labelformat=empty}
		\captionsetup{justification=centering}
		\caption*{Target  \\ \quad}
	\end{minipage}
\vskip-10pt
	\caption{Sample results compared with  state-of-the-art methods evaluated on \emph{Protocol1}.}  \label{fig:v1}
\end{figure*}
\begin{figure}[h]
\centering
\includegraphics[width=0.5\textwidth]{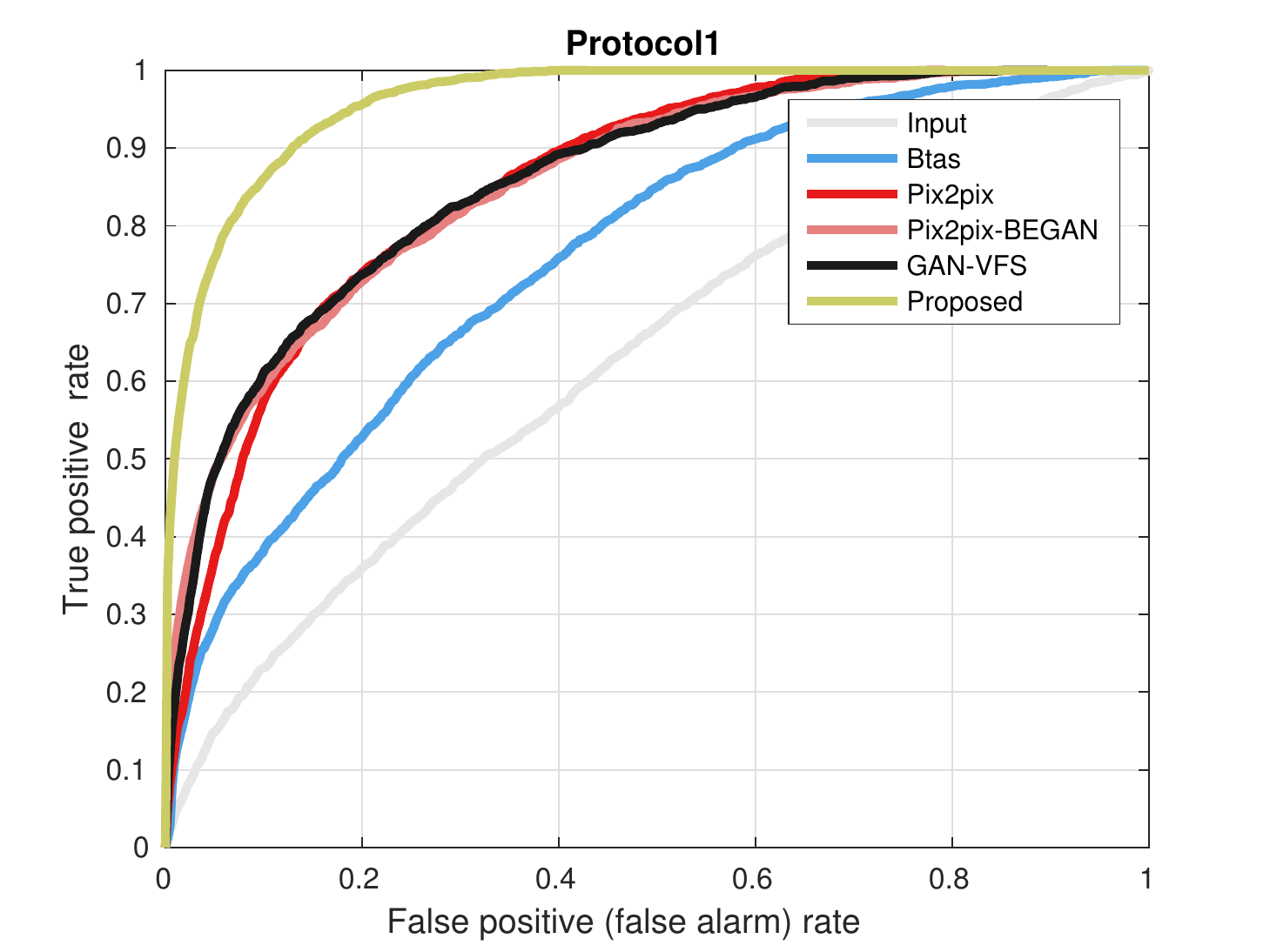}
 \vskip -6pt  \caption{The ROC curves corresponding to \emph{Protocol1}. }
\label{fig:test1}
\end{figure}

In addition to comparing the performance of the synthesized images in terms of SSIM and PSNR, we also compare the contribution of each module in face verification by plotting the ROC curves.  The verification results are evaluated based on the cosine similarity using the deep features extracted from the pre-defined VGG-face model \cite{vggface}. The results are shown in Figure~\ref{fig:ab1}. From the ROC curves, it can be clearly observed that the proposed dense-residual network with  multi-level pooling can also provide some discriminative information. Similar results can also be observed from the EER  and AUC comparisons, tabulated in Table~\ref{ta:ab1_result_psnr}.\\ 
\setlength{\tabcolsep}{4pt}
\begin{table}[h]
\centering
\caption{The PSNR, 	SSIM and EER and AUC results corresponding to \emph{Protocol1}.}
\label{tab:test1_psnr}
\resizebox{.50\textwidth}{!}{%
\begin{tabular}{|c|c|c|c|c|c|c|}
\hline
  & I-Polar & Btas-2016 \cite{face_btas2016} & Pix2pix \cite{pix2pix} & Pix2pix-BEGAN \cite{pix2pix,began} & GAN-VFS \cite{face_ijcb2017} & Proposed  \\ \hline
PSNR (dB) & 11.74 & 16.12 & 16.79 & 17.55 & 18.07 & \textbf{19.55}\\ \hline
SSIM & 0.4625 & 0.6785 & 0.6490 & 0.7033 & 0.7041 & \textbf{0.7433} \\ \hline
EER & 41.51\% & 26.72\% & 22.61\% & 22.56\% & 23.19\% & \textbf{11.78}\%\\ \hline
AUC & 62.93\% & 81.90\% & 85.14\% & 85.30\% & 85.89\%& \textbf{96.03}\% \\ \hline
\end{tabular}%
}
\end{table}
\setlength{\tabcolsep}{4pt}

\noindent {\bf{Ablation 2}}\\
The second ablation study is conducted to demonstrate the effectiveness of the proposed feature level multi-model fusion by conducting experiments with the following baselines:
\begin{enumerate}[nolistsep]
\item[(a)] \textbf{S-$S_0$:} Single stream dense-resisual encoder-decoder with the proposed structure with $S_0$ as the input. 
\item[(b)] \textbf{S-$S_1$:} Single stream dense-resisual encoder-decoder with the proposed structure with $S_1$ as the input
\item[(c)] \textbf{S-$S_2$:} Single stream dense-resisual encoder-decoder with the proposed structure with $S_2$ as the input.
\item[(d)] \textbf{S-Polar-IF:} Single stream dense-resisual encoder-decoder with the proposed structure with Polar as the input (i.e. input level fusion).  The S-Polar-IF model shares the exact same structure as DR-ED-ML as discussed in \textbf{Ablation 1}.
\item[(e)] \textbf{M-Polar-OF:} Multi stream dense-resisual encoder-decoder structure with output level fusion.  The M-Polar-OF is basically composed of three stream dense-resisual encoder-decoder structure, where each stream shares the same structure with S-Polar-IF but with different input ($S_0$, $S_1$ and $S_2$) for each stream. Then, the output features from each stream are fused (concatenated) at the end of the decoding part to generate visible face images.
\item[(f)] \textbf{M-Polar-F-L2:} Multi-stream dense-resisual encoder-decoder with the proposed structure based on feature-level fusion optimized with L2 loss only.
\item[(g)] \textbf{M-Polar-F-L2-GAN:} Multi-stream dense-resisual encoder-decoder with the proposed structure based on feature-level fusion optimized with L2 and GAN loss.
\item[(h)] \textbf{M-Polar-F-L2-GAN-Perp:} Multi-stream dense-resisual encoder-decoder with the proposed structure based on feature-level fusion optimized with L2, GAN loss and perceptual loss.
\item[(i)] \textbf{Our (M-Polar-FF):} Multi-stream dense-resisual encoder-decoder with the proposed structure based on feature-level fusion with all the losses.
\end{enumerate}

\begin{figure*}[t]
	\centering
		\begin{minipage}{.13\textwidth}
			\centering
			\caption*{PSNR: 10.36 \\ SSIM: 0.4331}\vskip-10pt
			\includegraphics[width=1\textwidth]{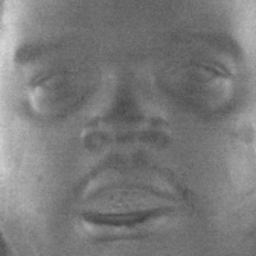}
			\captionsetup{labelformat=empty}
			\captionsetup{justification=centering}
	\end{minipage}
		\begin{minipage}{.13\textwidth}
			\centering
			\caption*{PSNR: 14.67 \\ SSIM: 0.6301}\vskip-10pt
			\includegraphics[width=1\textwidth]{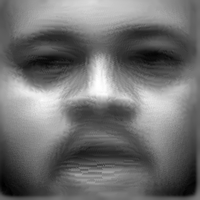}
			\captionsetup{labelformat=empty}
			\captionsetup{justification=centering}
	\end{minipage}
	\begin{minipage}{.13\textwidth}
		\centering
			\caption*{PSNR:15.74 \\ SSIM: 0.6305}\vskip-10pt
			\includegraphics[width=1\textwidth]{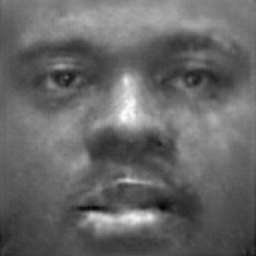}
		\captionsetup{labelformat=empty}
		\captionsetup{justification=centering}
	\end{minipage}
	\begin{minipage}{.13\textwidth}
		\centering
			\caption*{PSNR:20.04 \\ SSIM: 0.7004}\vskip-10pt
			\includegraphics[width=1\textwidth]{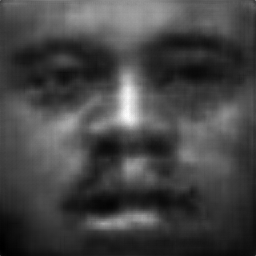}
		\captionsetup{labelformat=empty}
		\captionsetup{justification=centering}
	\end{minipage}
	\begin{minipage}{.13\textwidth}
		\centering
			\caption*{PSNR:19.86 \\ SSIM: 0.7249  }\vskip-10pt
			\includegraphics[width=1\textwidth]{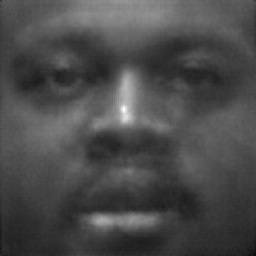}
		\captionsetup{labelformat=empty}
		\captionsetup{justification=centering}
	\end{minipage}
	\begin{minipage}{.13\textwidth}
		\centering
			\caption*{PSNR: \textbf{21.66} \\ SSIM: \textbf{0.7728}}\vskip-10pt
			\includegraphics[width=1\textwidth]{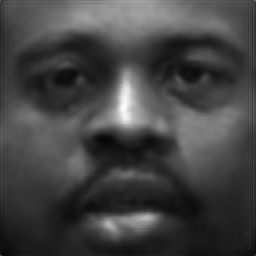}
		\captionsetup{labelformat=empty}
		\captionsetup{justification=centering}
	\end{minipage}
	\begin{minipage}{.13\textwidth}
		\centering
			\caption*{PSNR:Inf \\ SSIM: 1.0000}
\vskip-10pt
			\includegraphics[width=1\textwidth]{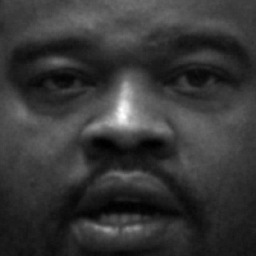}
		\captionsetup{labelformat=empty}
		\captionsetup{justification=centering}
	\end{minipage}\\	\vskip+5pt
		\begin{minipage}{.13\textwidth}
			\centering
			\caption*{PSNR: 10.25 \\ SSIM: 0.4108}\vskip-10pt
			\includegraphics[width=1\textwidth]{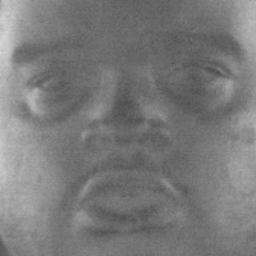}
			\captionsetup{labelformat=empty}
			\captionsetup{justification=centering}
	\end{minipage}
		\begin{minipage}{.13\textwidth}
			\centering
			\caption*{PSNR: 14.67 \\ SSIM: 0.6301}\vskip-10pt
			\includegraphics[width=1\textwidth]{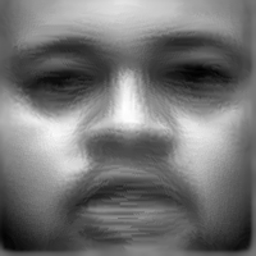}
			\captionsetup{labelformat=empty}
			\captionsetup{justification=centering}
	\end{minipage}
	\begin{minipage}{.13\textwidth}
		\centering
			\caption*{PSNR:13.98 \\ SSIM: 0.5915}\vskip-10pt
			\includegraphics[width=1\textwidth]{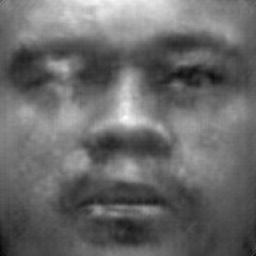}
		\captionsetup{labelformat=empty}
		\captionsetup{justification=centering}
	\end{minipage}
	\begin{minipage}{.13\textwidth}
		\centering
			\caption*{PSNR:20.04 \\ SSIM: 0.7544}\vskip-10pt
			\includegraphics[width=1\textwidth]{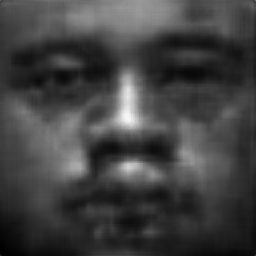}
		\captionsetup{labelformat=empty}
		\captionsetup{justification=centering}
	\end{minipage}
	\begin{minipage}{.13\textwidth}
		\centering
			\caption*{PSNR:19.67 \\ SSIM: 0.7306  }\vskip-10pt
			\includegraphics[width=1\textwidth]{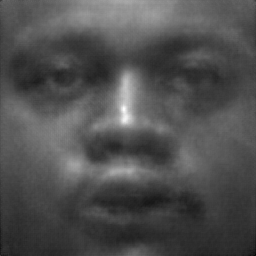}
		\captionsetup{labelformat=empty}
		\captionsetup{justification=centering}
	\end{minipage}
	\begin{minipage}{.13\textwidth}
		\centering
			\caption*{PSNR: \textbf{23.55} \\ SSIM: \textbf{0.8097}}\vskip-10pt
			\includegraphics[width=1\textwidth]{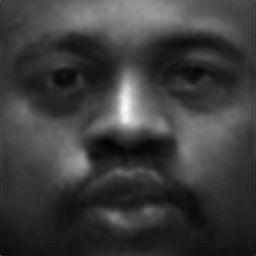}
		\captionsetup{labelformat=empty}
		\captionsetup{justification=centering}
	\end{minipage}
	\begin{minipage}{.13\textwidth}
		\centering
			\caption*{PSNR:Inf \\ SSIM: 1.0000}
\vskip-10pt
			\includegraphics[width=1\textwidth]{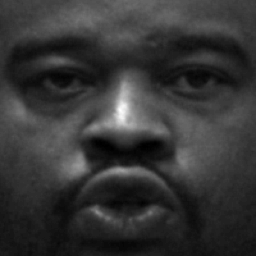}
		\captionsetup{labelformat=empty}
		\captionsetup{justification=centering}
	\end{minipage}\\	\vskip+5pt
		\begin{minipage}{.13\textwidth}
			\centering
			\caption*{PSNR: 11.01 \\ SSIM: 0.4236}\vskip-10pt
			\includegraphics[width=1\textwidth]{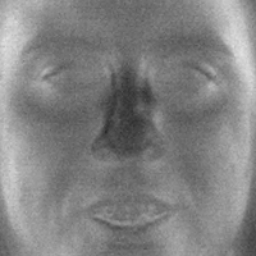}
			\captionsetup{labelformat=empty}
			\captionsetup{justification=centering}
	\end{minipage}
		\begin{minipage}{.13\textwidth}
			\centering
			\caption*{PSNR: 18.99 \\ SSIM: 0.7525}\vskip-10pt
			\includegraphics[width=1\textwidth]{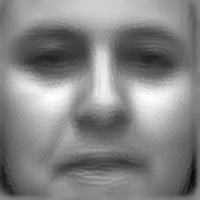}
			\captionsetup{labelformat=empty}
			\captionsetup{justification=centering}
	\end{minipage}
	\begin{minipage}{.13\textwidth}
		\centering
			\caption*{PSNR:22.96 \\ SSIM: 0.8008}\vskip-10pt
			\includegraphics[width=1\textwidth]{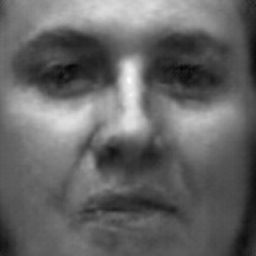}
		\captionsetup{labelformat=empty}
		\captionsetup{justification=centering}
	\end{minipage}
	\begin{minipage}{.13\textwidth}
		\centering
			\caption*{PSNR:23.92 \\ SSIM: 0.8421}\vskip-10pt
			\includegraphics[width=1\textwidth]{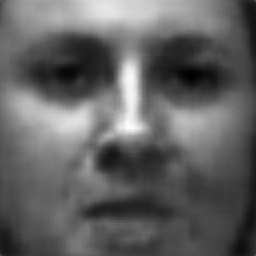}
		\captionsetup{labelformat=empty}
		\captionsetup{justification=centering}
	\end{minipage}
	\begin{minipage}{.13\textwidth}
		\centering
			\caption*{PSNR:25.36 \\ SSIM: 0.8572  }\vskip-10pt
			\includegraphics[width=1\textwidth]{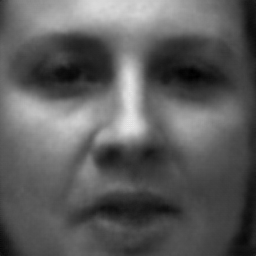}
		\captionsetup{labelformat=empty}
		\captionsetup{justification=centering}
	\end{minipage}
	\begin{minipage}{.13\textwidth}
		\centering
			\caption*{PSNR:\textbf{26.32} \\ SSIM: \textbf{0.8732}}\vskip-10pt
			\includegraphics[width=1\textwidth]{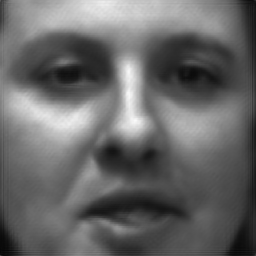}
		\captionsetup{labelformat=empty}
		\captionsetup{justification=centering}
	\end{minipage}
	\begin{minipage}{.13\textwidth}
		\centering
			\caption*{PSNR:Inf \\ SSIM: 1.0000}
\vskip-10pt
			\includegraphics[width=1\textwidth]{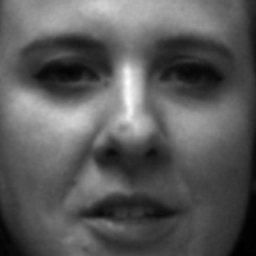}
		\captionsetup{labelformat=empty}
		\captionsetup{justification=centering}
	\end{minipage} \vskip+5pt
		\begin{minipage}{.13\textwidth}
			\centering
			\caption*{PSNR: 9.13 \\ SSIM: 0.3743}\vskip-10pt
			\includegraphics[width=1\textwidth]{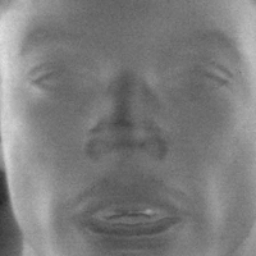}
			\captionsetup{labelformat=empty}
			\captionsetup{justification=centering}
		\caption*{I-Polar  \\ \quad}
	\end{minipage}
		\begin{minipage}{.13\textwidth}
			\centering
			\caption*{PSNR: 14.36 \\ SSIM: 0.6052}\vskip-10pt
			\includegraphics[width=1\textwidth]{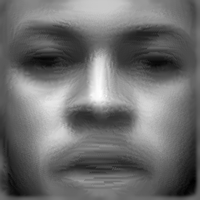}
			\captionsetup{labelformat=empty}
			\captionsetup{justification=centering}
		\caption*{Btas-2016 \\ \cite{face_btas2016}}
	\end{minipage}
	\begin{minipage}{.13\textwidth}
		\centering
			\caption*{PSNR:14.79 \\ SSIM: 0.6087}\vskip-10pt
			\includegraphics[width=1\textwidth]{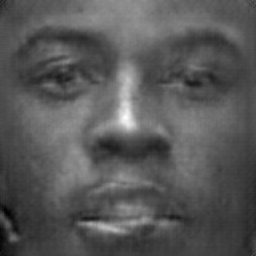}
		\captionsetup{labelformat=empty}
		\captionsetup{justification=centering}
		\caption*{Pix2pix \\ \cite{pix2pix}}
	\end{minipage}
	\begin{minipage}{.13\textwidth}
		\centering
			\caption*{PSNR:17.48 \\ SSIM: 0.7015}\vskip-10pt
			\includegraphics[width=1\textwidth]{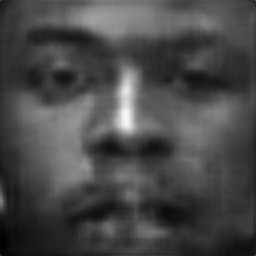}
		\captionsetup{labelformat=empty}
		\captionsetup{justification=centering}
		\caption*{Pix2pix-BEGAN \\ \cite{pix2pix,began}  }
	\end{minipage}
	\begin{minipage}{.13\textwidth}
		\centering
			\caption*{PSNR:17.69 \\ SSIM: 0.7019  }\vskip-10pt
			\includegraphics[width=1\textwidth]{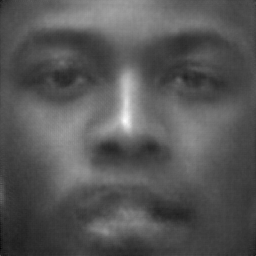}
		\captionsetup{labelformat=empty}
		\captionsetup{justification=centering}
		\caption*{GAN-VFS \\ \cite{face_ijcb2017}}
	\end{minipage}
	\begin{minipage}{.13\textwidth}
		\centering
			\caption*{PSNR:\textbf{22.45} \\ SSIM: \textbf{0.8107}}\vskip-10pt
			\includegraphics[width=1\textwidth]{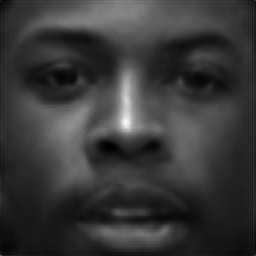}
		\captionsetup{labelformat=empty}
		\captionsetup{justification=centering}
		\caption*{Proposed \\ \quad}
	\end{minipage}
	\begin{minipage}{.13\textwidth}
		\centering
			\caption*{PSNR:Inf \\ SSIM: 1.0000}
\vskip-10pt
			\includegraphics[width=1\textwidth]{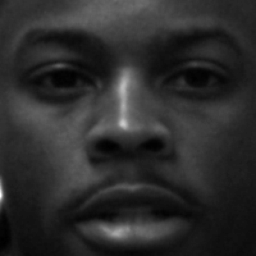}
		\captionsetup{labelformat=empty}
		\captionsetup{justification=centering}
		\caption*{Target \\ \qquad}
	\end{minipage}\\	\vskip-10pt
	\caption{Sample results compared with  state-of-the-art methods evaluated on \emph{Protocol2}.}  \label{fig:v2}
\end{figure*}
\begin{figure}[h]
\centering
\includegraphics[width=0.5\textwidth]{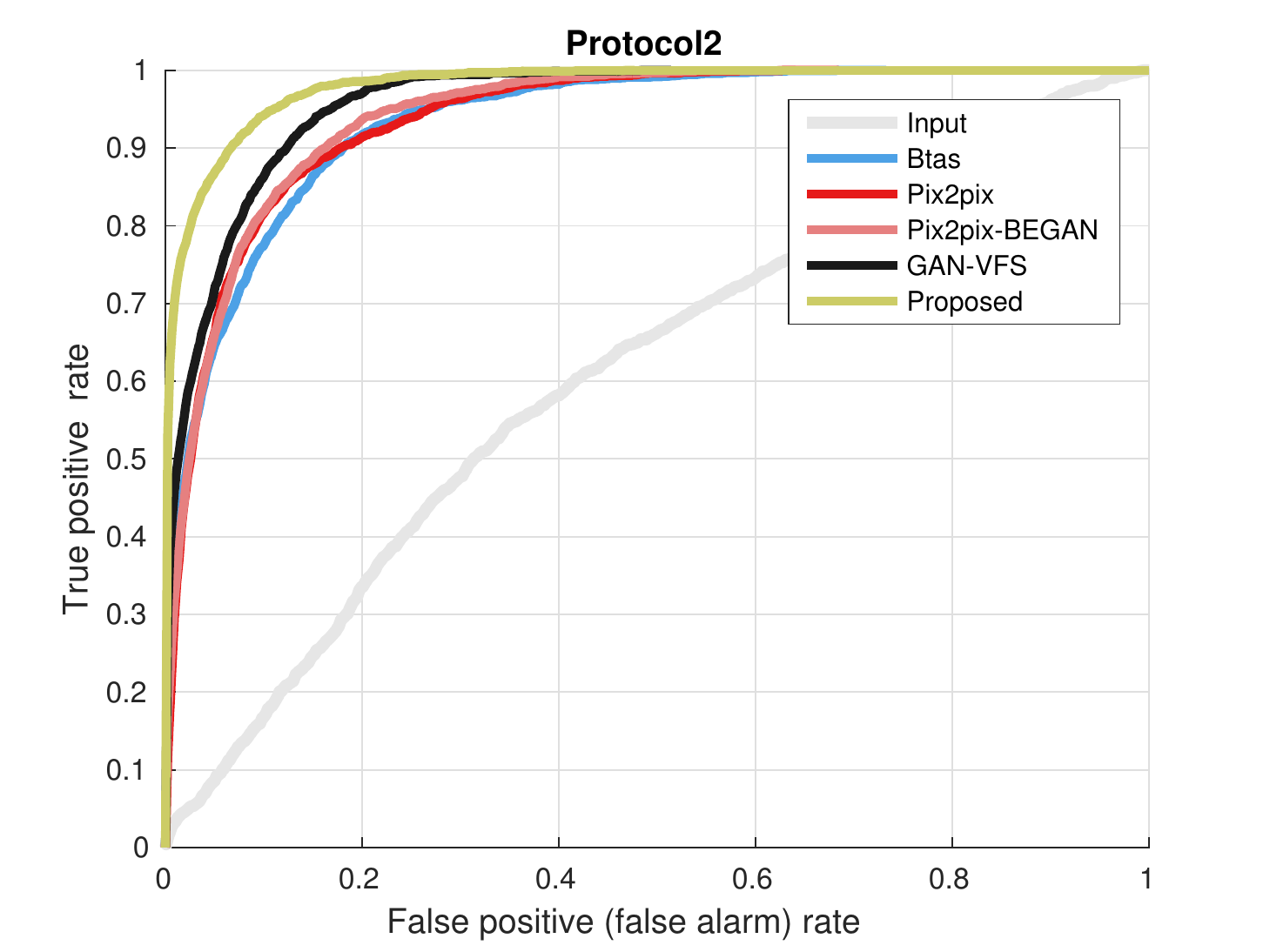}
 \vskip -6pt  \caption{The ROC curves corresponding to the \emph{
 Protocol2}. }
\label{fig:test2}
\end{figure}
Sample results corresponding to \textbf{Ablation 2} is shown in Figure \ref{fig:ablation2}. It can be observed that just leveraging any one of the Stokes images as input is unable to fully capture the geometric and texture details of the whole face. For example, as shown in the first column second row in Figure \ref{fig:ablation2},  the nose is over-synthesized if just $S_0$ (representing conventional thermal imagery) is used.  Leveraging input level fusion (just concatenating three modalities as three-channel input) S-Polar-IF enables better visible face with less undesired artifacts as compared to S-$S_0$, S-$S_1$ and S-$S_2$.   Furthermore, the proposed multi-stream feature-level fusion structure is able to preserve more geometric facial details and is able to generate photo-realistic visible face images. Visual results also demonstrate the effectiveness of leveraging feature level fusion over input level or output level fusion.  Quantitative results evaluated in terms of PSNR and SSIM are shown in Table~\ref{tab:ablation2}.  Results are also consistent with our visual comparison.

Similar to Ablation study 1, the face verification results are also used as a metric to evaluate the performnace of different fusion techniques.  We plot the ROC curves corresponding to the different settings discussed above.    The ROC curves are shown in Figure~\ref{fig:ab2}. Again, the verification results are evaluated  based the cosine similarity using the deep features extracted from the VGG-face model \cite{vggface} without fine-tuning.  From the ROC curves, it can be clearly observed that the proposed multi-stream feature-level fusion  can bring in more discriminative information as compared to  input level or output level fusion.\\  
\begin{table}[h]
\centering
\caption{The PSNR, 	SSIM, EER and AUC results corresponding to \emph{Protocol2}.}
\label{tab:test2_psnr}
\resizebox{.5\textwidth}{!}{%
\begin{tabular}{|c|c|c|c|c|c|c|}
\hline
  & I-Polar & Btas-2016 \cite{face_btas2016} & Pix2pix \cite{pix2pix} & Pix2pix-BEGAN \cite{pix2pix,began} & GAN-VFS \cite{face_ijcb2017} & Proposed  \\ \hline
PSNR (dB) & 10.88 & 15.82 & 17.82 & 18.28 & 18.58 & \textbf{19.18}\\ \hline
SSIM & 0.4467 & 0.6854 & 0.6828 & 0.7214 & 0.7283& \textbf{0.7340}\ \\ \hline
EER & 40.87\% & 14.60\% & 13.49\% & 15.81\% & 11.42\% & \textbf{7.99}\%\\ \hline
AUC & 61.27\% & 93.99\% & 93.46\% & 92.50\% & 95.96\%& \textbf{98.00}\% \\ \hline
\end{tabular}%
}
\end{table}

\noindent {\bf{Ablation 3}}\\
In the third ablation study, we demonstrate the effectiveness of different loss functions used in the proposed method (e.g. adversarial loss,  perceptual loss and identity preserving loss) by conducting the following experiments. All the experimental results are evaluated using  \textbf{Protocol 1} based on the polarimetric images as the input:  
\begin{enumerate}[nolistsep]
	\item[(a)] \textbf{L2}: The proposed architecture (M-Polar-FF) optimized with the L2 loss. 
	\item[(b)] \textbf{L2-GAN}: The proposed architecture optimized with the L2 loss and the adversarial loss.
	\item[(c)] \textbf{L2-GAN-P}: The proposed architecture optimized with the L2 loss, the adversarial loss and the perceptual loss.
	\item[(d)] \textbf{Our}: The proposed architecture optimized with the L2 loss, the adversarial loss, the perceptual loss and the identity-preserving loss.
\end{enumerate}
Visual results corresponding to this ablation study are shown in Figure~\ref{fig:ab3}. It can be observed from the results that the L2 loss itself generates blurry faces and many details around the eyes and the mouth regions are missing. By involving the GAN structure in the proposed method, more details are being added to the results.  But it can be observed that GAN itself produces images with artifacts. Introduction of the perceptual loss in the proposed  framework  is  able  to  remove some of the artifacts  and makes the results visually pleasing.  Finally, the combination of all the losses is able to generate more reasonable results with better facial details.

To better demonstrate the effectiveness of different losses in the proposed method, we plot the ROC curves corresponding to the above four different network settings.   The results are shown in Figure~\ref{fig:ab32}. All the verification results are evaluated on the deep features extracted from the VGG-face model \cite{vggface} without fine-tuning.  From the ROC curves, it can be clearly observed that even though the identity loss does not produce visually different results, it can bring in more discriminative information. The corresponding PSNR, SSIM values as well as the AUC and EER values are summarized in  Table~\ref{ta:ab3_result_psnr}.

\subsection{Comparison with State-of-the-Art Methods}
To demonstrate the improvements achieved by the proposed method, it is compared against recent state-of-the-art
methods \cite{face_btas2016,pix2pix,began,face_ijcb2017} on the new dataset.  We compare quantitative and qualitative performance of different methods on the test images from the two distinct protocols  \emph{Protocol1} and \emph{Protocol2} discussed earlier.

Sample results corresponding to Protocol 1 and Protocol 2 are shown in Figure~\ref{fig:v1} and Figure~\ref{fig:v2}, respectively. It can be observed from these figures, Pix2pix and Pix2pix-BEGAN introduce undesirable artifacts in the final reconstructed images.  

The introduction of the perceptual loss in \cite{face_ijcb2017} is able to remove some of these artifacts and produce visually pleasing results.  However, the synthesized images still lack some geometric and texture details as compared to the target image. In contrast, the proposed method is able to generate photo-realistic visible face images while better retaining the discriminative information such as the structure of mouth and eye. Quantitative results corresponding to different methods evaluated on both protocols are tabulated in Table~\ref{tab:test1_psnr} and Table\ref{tab:test2_psnr}, showing  that the proposed multi-stream feature-level fusion GAN structure is able to achieve superior performance. 

Similar to the ablation study, we also propose to use the performance of face verification as a metric to evaluate the performance of different methods.  Figure\ref{fig:test1} and Figure\ref{fig:test2} show the ROC curves corresponding to the two experimental protocols. The AUC and EER results are reported in Table~\ref{tab:test1_psnr} and Table~\ref{tab:test2_psnr}. From these results, it  can be clearly observed that the proposed method is able to achieve superior  quantitative  performance  compared  the  previous approaches.  These results highlight the significance of using a GAN-based approach to image synthesis.

\section{Conclusion}
\label{sec:con}
We present a new multi-level dense-residual fusion GAN structure for synthesizing photo-realistic visible face images from the corresponding polarimetric data. In contrast to the previous methods that leverage input level fusion techniques to combine geometric and texture information from different Stokes image, we take a different approach where  visual features extracted from different Stokes images  are combined to synthesize the photo-realistic face images. Quantitative and qualitative experiments evaluated on a real polarimetric visible database demonstrate that the proposed method is able to achieve significantly better results as compared to the recent state-of-the-art methods. In addition, three ablation studies are performed to demonstrate the improvements obtained by the feature-level fusion methods, different modules and different loss functions in the proposed method. Furthermore, an extended polarimetric-visible database consisting of data from 111 subjects is also presented in this paper.

\begin{acknowledgements}
 We  like  to  thank  Vishwanath A. Sindagi,  for  his
insightful discussion on this topic.
\end{acknowledgements}

\bibliographystyle{spmpsci}      
\bibliography{egbib}   

\end{document}